\documentclass[10pt,twocolumn,letterpaper]{article}

\usepackage{cvpr}              % To produce the CAMERA-READY version
\usepackage[dvipsnames]{xcolor}

\definecolor{cvprblue}{rgb}{0.21,0.49,0.74}
\usepackage[pagebackref,breaklinks,colorlinks,citecolor=cvprblue]{hyperref}
\usepackage{multirow}

\usepackage{colortbl}
\usepackage{float} 

\usepackage{soul} % for highlighting
\usepackage{xcolor} % for custom colors

\sethlcolor{cyan} % for base categories
\sethlcolor{yellow} % for novel categories
\definecolor{dpurple}{RGB}{128, 0, 128}
\definecolor{brown}{RGB}{255, 192, 203}

\def\twod{\text{2D}}
\def\threed{\text{3D}}
\def\method{\textsc{OVMono3D}\xspace}

\def\paperID{107} % *** Enter the Paper ID here
\def\confName{3DV\xspace}
\def\confYear{2026\xspace}

\title{Open Vocabulary Monocular 3D Object Detection}

\author{Jin Yao$^{1}$ \quad Hao Gu$^1$ \quad Xuweiyi Chen$^1$ \quad Jiayun Wang$^2$ \quad Zezhou Cheng$^1$ \\
$^1$ University of Virginia \quad $^2$ California Institute of Technology\\
\url{https://uva-computer-vision-lab.github.io/ovmono3d/}
} 

\begin{document}

\makeatletter
\g@addto@macro\@maketitle{
  % \vspace{-6mm}
  \begin{figure}[H]
  \setlength{\linewidth}{\textwidth}
  \setlength{\hsize}{\textwidth}
  \centering
  \includegraphics[width=\linewidth]{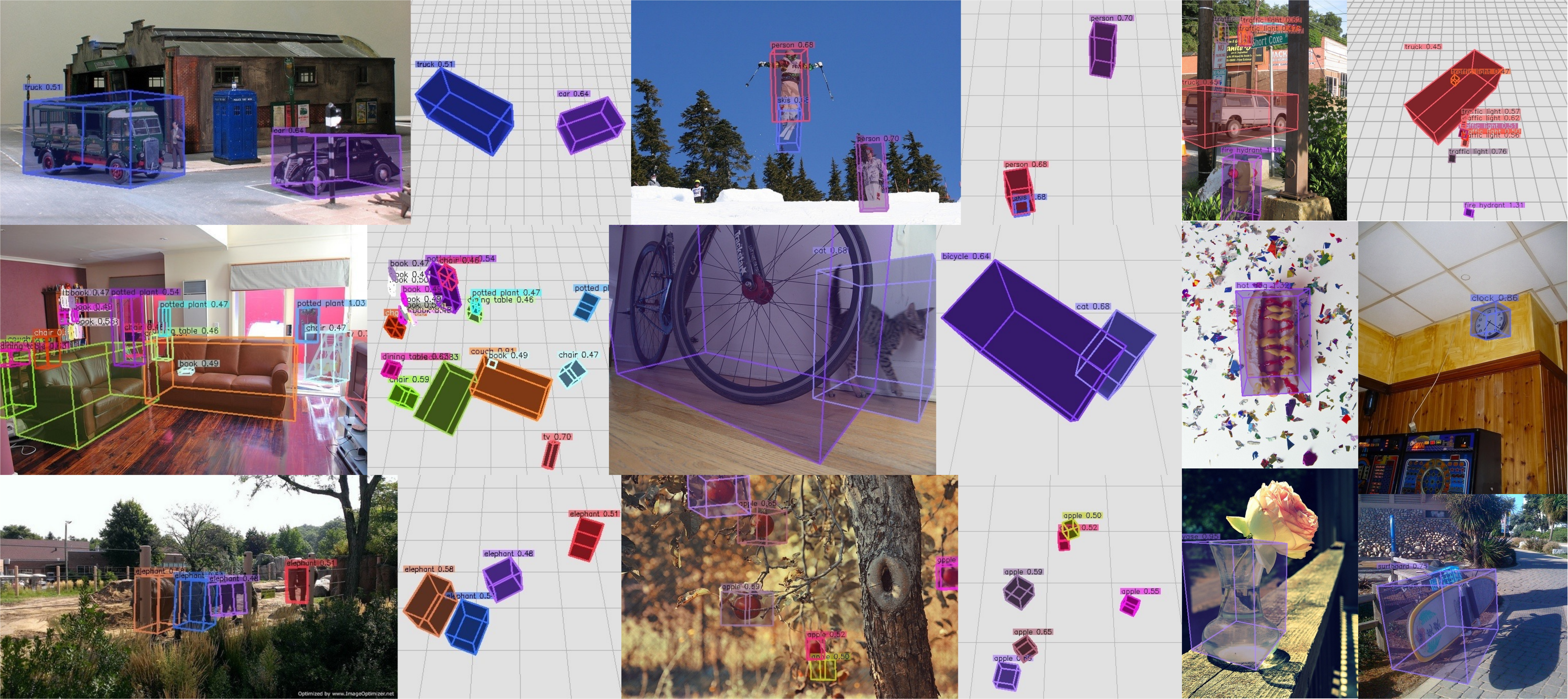}
  % \vspace{-2mm}
  \caption{\textbf{Our Model on In-the-Wild COCO~\cite{lin2014microsoft} Images}. We display 3D predictions overlaid on the images and the top-down views with a base grid of $1\,\text{m} \times 1\,\text{m}$ tiles. For single-object images, only front-views are displayed.
    Results are shown for both base and novel categories, demonstrating that our proposed method exhibits zero-shot generalization capability on real-world images.}
  \label{fig:coco}
  \end{figure}
}

\maketitle

\begin{abstract}

We propose and study open-vocabulary monocular 3D detection, a novel task that aims to detect objects of any categories in metric 3D space from a single RGB image. 
Existing 3D object detectors either rely on costly sensors such as LiDAR or multi-view setups, or remain confined to closed vocabularies settings with limited categories, restricting their applicability.  
We identify two key challenges in this new setting. 
First, the scarcity of 3D bounding box annotations limits the ability to train generalizable models. To reduce dependence on 3D supervision, we propose a framework that effectively integrates pretrained 2D and 3D vision foundation models. Second, missing labels and semantic ambiguities (\eg, table vs. desk) in existing datasets hinder reliable evaluation. To address this, we design a novel metric that captures model performance while mitigating annotation issues. Our approach achieves state-of-the-art results in zero-shot 3D detection of novel categories as well as in-domain detection on seen classes. We hope our method provides a strong baseline and our evaluation protocol establishes a reliable benchmark for future research.

\end{abstract}

% \vspace{-3mm}
\section{Introduction}
\label{sec:intro}

Recognizing objects from a single image has been a longstanding task in computer vision, with broad applications in robotics and AR/VR. 
Over recent decades, 2D object detection --- identifying and localizing objects within a 2D image plane --- has achieved substantial progress, driven by advancements in deep learning techniques~\cite{girshick2015fast,ren2016faster,he2017mask,carion2020detr} and large annotated datasets~\cite{lin2014microsoft,gupta2019lvis, shao2019objects365}.
However, 
recognizing only a fixed set of objects is limiting, given the vast diversity of objects in real-world settings; 
detecting objects solely in 2D space is also insufficient for most practical tasks, as the world and its objects exist in 3D space.

\begin{figure}[!t]
    \centering
    % \vspace{-3mm}
    \includegraphics[width=\columnwidth]{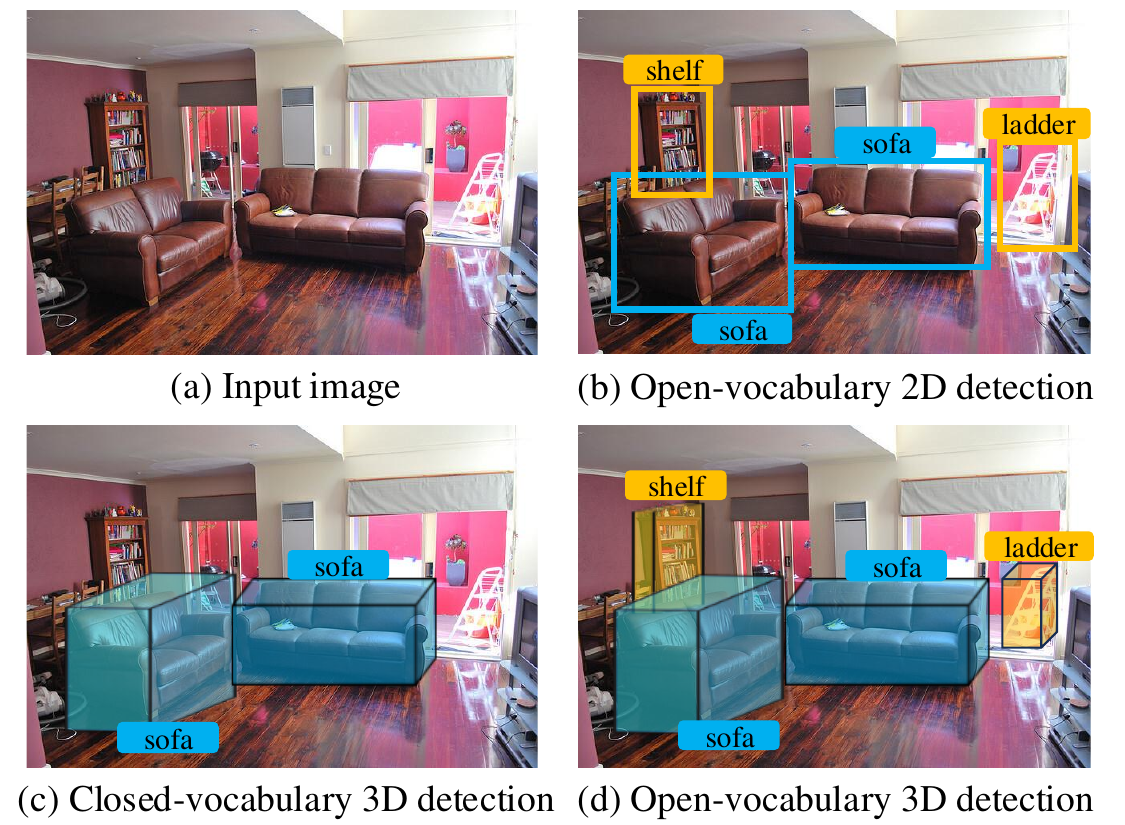}
    % \vspace{-6mm}
\caption{Given \textbf{(a)} a single image, we illustrate examples of \textbf{(b)} open-vocabulary 2D detection, which localizes objects of any category within the 2D image plane, covering both \sethlcolor{cyan}\hl{\textbf{seen}} categories and \sethlcolor{yellow}\hl{\textbf{novel}} categories not seen during training; \textbf{(c)} closed-vocabulary 3D detection, which detects objects from a predefined set of categories in 3D space; and \textbf{(d)} open-vocabulary 3D object detection, which identifies objects of any category in 3D.}
\label{fig:teaser}
\end{figure}

To address these limitations, one line of recent research has focused on open-vocabulary 2D object detection~\cite{liu2023grounding,cheng2024yolo,wu2024clim,zhou2022extract,wu2023open} (Fig.~\ref{fig:teaser}b) to identify objects beyond a fixed set of categories. 
Another line explores the monocular 3D detection task~\cite{brazil2023omni3d, rukhovich2022imvoxelnet,wang2022probabilistic, nie2020total3dunderstandingjointlayoutobject} (Fig.~\ref{fig:teaser}c), which extends detection capabilities from 2D to 3D space. 
Despite the vast research in these two lines of research, 
the intersection of these two areas --- open-vocabulary monocular 3D detection, referred to as \method{} (Fig.~\ref{fig:teaser}d) --- remains largely unexplored. 
In this work, we aim to fill this gap. The \method{} task involves detecting and localizing objects of any categories in the metric 3D space, including novel categories unseen during training. 

\begin{figure}[t!]
    \centering
    \includegraphics[width=\linewidth]{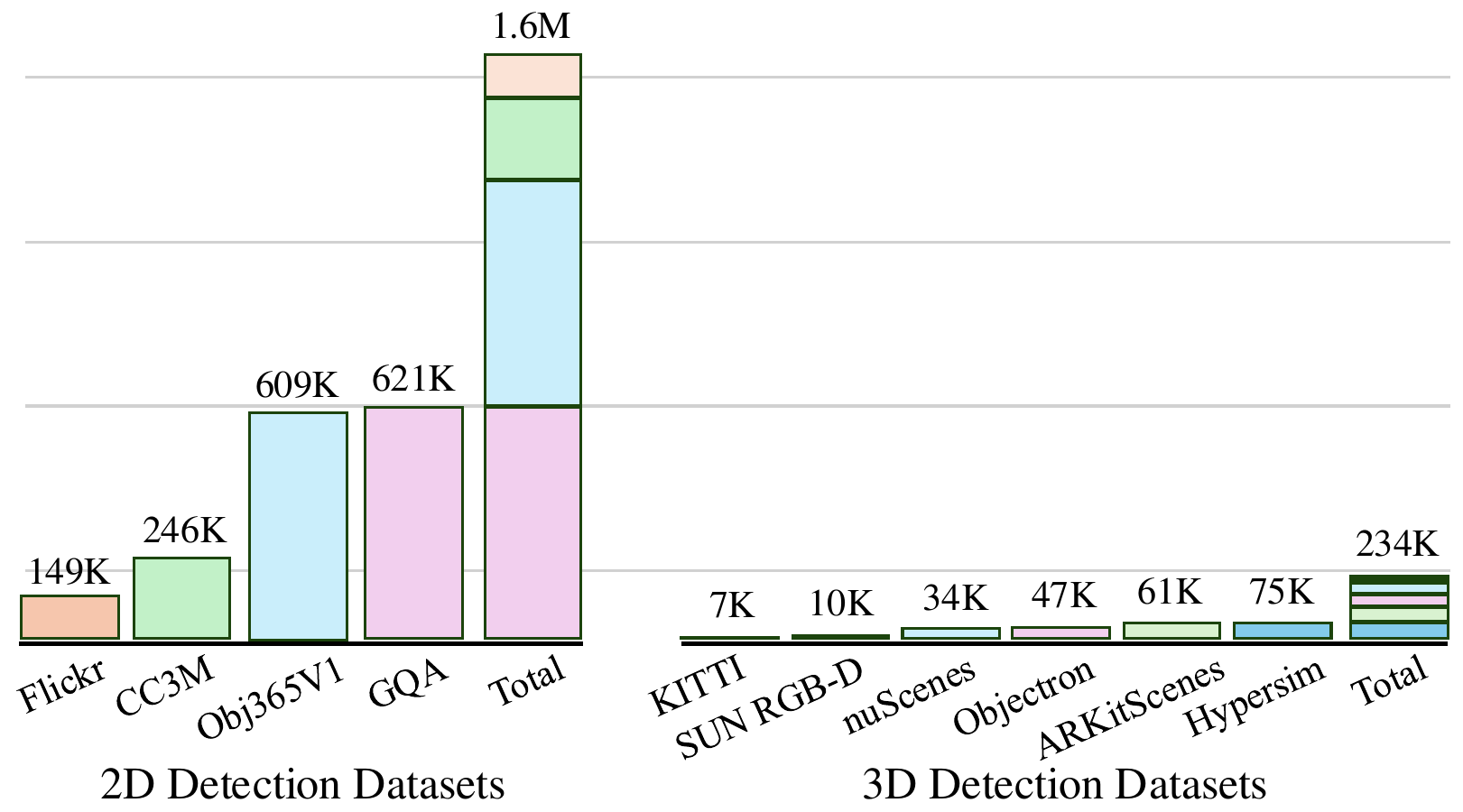}
     \caption{\textbf{2D vs. 3D Detection Datasets in terms of \#Images.} Publicly available 3D datasets with 3D annotations are significantly smaller than 2D detection datasets.}
    \label{fig:3d-dataset}
    % \vspace{-5mm}
\end{figure}

One of the central challenges in \method{} lies in \textbf{generalizable model training}, as large-scale 3D datasets with high-quality annotations are scarce. 
Unlike open-vocabulary 2D detection, where abundant training data exists, 3D datasets are limited in scale (Fig.~\ref{fig:3d-dataset}). 
To address this, we adopt a decoupled strategy that separates 2D recognition and localization from 3D bounding box estimation. 
Specifically, we explore two complementary approaches:
(1) \textbf{\method{}-GEO}, a simple training-free baseline that unprojects 2D detections from off-the-shelf open-vocabulary detectors~\cite{cheng2024yolo,liu2023grounding} into 3D using geometric principles (Fig.~\ref{fig:method}a). While requiring no 3D annotations and achieving reasonable performance, this approach is sensitive to occlusion and degrades in cluttered scenes;
(2) \textbf{\method{}-LIFT}, a data-driven method that integrates multiple vision foundation models (\eg, depth estimation~\cite{piccinelli2024unidepth}, 2D open-vocabulary detection~\cite{cheng2024yolo,liu2023grounding}, and general image encoder~\cite{oquab2023dinov2}) and learns to lift 2D detections into 3D (Fig.~\ref{fig:method}b). 
We systematically analyze its design space, including backbone architectures, 2D base detectors, and other critical choices (Sec.~\ref{sec:analysis}).
Our results highlight the importance of 3D-aware image representations and accurate depth estimation for robust performance. 
Compared to the geometric baseline, \method{}-LIFT shows significantly greater resilience to occlusion and achieves superior results across diverse, real-world scenarios.

Another central challenge is \textbf{reliable model evaluation}, hindered by missing annotations and semantic ambiguity. 
The task requires detecting all visible objects in 2D images, yet existing 3D datasets generally provide only partial annotations due to the high cost of labeling. 
For example, in SUN RGB-D~\cite{song2015sun}, a chair visible in the RGB image may be unannotated in 3D if it was partially occluded in the depth scan, leading to false negatives in evaluation. 
Similarly, inconsistencies in naming conventions across datasets, such as predicting desk when the ground truth label is table, cause correct detections to be miscounted as false positives under standard metrics. 
To address these issues, we propose a simple evaluation protocol that reduces the impact of missing labels and semantic variation (Sec.~\ref{subsec:evaluation_prototype}). 
The resulting metric enables reliable assessment without exhaustive annotations and is particularly suited for in-the-wild images, where complete labeling is infeasible and error-prone.

Finally, we conduct extensive experiments and ablations to validate both our proposed approaches and evaluation metric on standard benchmarks. 
Our method not only surpasses concurrent open-vocabulary 3D detectors (\eg, DetAny3D~\cite{zhang2025detect}) by a large margin (Tab.~\ref{tab:main}), but also outperforms closed-vocabulary counterparts (\eg, Cube R-CNN\cite{brazil2023omni3d}, UniMODE~\cite{li2024unified3dobjectdetection}) on in-domain categories (Tab.~\ref{tab:closed_vocab_result}), under both standard and proposed metrics. 
Qualitative results further highlight strong generalization to in-the-wild images (Fig.~\ref{fig:coco}). 
We hope these insights inspire further exploration of advanced architectures for this task. Code will be released to support future research.

\section{Related Work}
\label{sec:related}
\paragraph{Open-Vocabulary 2D Object Detection.} 
The goal of this task is to recognize and localize objects in 2D images beyond a fixed set of predefined categories. Leveraging large-scale 2D datasets~\cite{shao2019objects365, kamath2021mdetr, ordonez2011im2text}, this field has seen considerable advancement. Some approaches~\cite{zareian2021open, gu2021open, minderer2022simple, ma2022open} employ pre-trained vision-language models~\cite{jia2021scaling,radford2021learning}, using frozen text features to detect novel categories. Other methods~\cite{li2022grounded, zhang2022glipv2, liu2023grounding, cheng2024yolo} are pre-trained on extensive detection, grounding, and caption data to align region-text features. For example, Grounding DINO~\cite{liu2023grounding} incorporates grounded pre-training with cross-modal feature fusion, while YOLO-World~\cite{cheng2024yolo} uses region-text contrastive loss and re-parameterization to enhance both accuracy and efficiency. However, such large-scale annotations are costly in 3D detection; thus, we investigate methods to adapt existing open-vocabulary 2D detectors for 3D detection.

\paragraph{Open-Vocabulary 3D Object Detection.} 
% This task aims at identifying objects of any categories in 3D.
This task seeks to identify objects from any category in 3D, including those unseen during training.
Prior research~\cite{lu2022open, lu2023open, cao2024coda, cao2024collaborative,zhang2024fm,zhu2023object2scene,zhang2022pointclip, zhang2025opensight, peng2025global, xia2024openad, peng2025glrd, Djamahl2024find, yao2025labelany3d} primarily focuses on 3D detection tasks with 3D point clouds as input. \cite{lu2022open} first proposes an open-vocabulary 3D detector using image-level class supervision from ImageNet1K~\cite{russakovsky2015imagenet}. \cite{lu2023open} utilizes 2D bounding boxes from a pre-trained 2D detector~\cite{zhou2022detecting} to build pseudo-3D label. \cite{zhang2024fm} leverages various 2D foundation models to enhance the performance of 3D open-vocabulary detection.
OV-Uni3DETR~\cite{wang2024ov} proposes a multi-modal open-vocabulary 3D detector that accommodates both point clouds and images as the input. 

In contrast, our work focuses on the monocular 3D detection task that only requires RGB images as input and does not assume the availability of point cloud data at training or inference phase.

\paragraph{Monocular 3D Object Detection} refers to the task of identifying and localizing objects within a scene using 3D bounding boxes derived from single-view images. Early research in this domain primarily targeted either outdoor~\cite{Chen_2016_CVPR_Mono_Auto_Driving, zhou2021iafainstanceawarefeatureaggregation, zhou2019objectspoints, wang2021fcos3dfullyconvolutionalonestage, chen2021monorunmonocular3dobject, zhang2023monodetr,huang2022monodtr, wang2022probabilistic} or indoor~\cite{Dasgupta_2016_CVPR, huang2019cooperativeholisticsceneunderstanding, nie2020total3dunderstandingjointlayoutobject,rukhovich2021imvoxelnetimagevoxelsprojection,lazarow2024cubify} environments, focusing on specialized applications such as autonomous driving and room layout estimation. The extensive Omni3D~\cite{brazil2023omni3d} dataset enabled Cube R-CNN~\cite{brazil2023omni3d} and UniMODE~\cite{li2024unified3dobjectdetection} to perform unified monocular 3D object detection across multiple scene types. 
Recent works also explore pseudo annotation generation to expand detector vocabularies. OVM3D-Det~\cite{huang2024training} generates pseudo labels enabling detection of novel categories without human annotations. V-MIND~\cite{jhang2025v} converts large-scale 2D datasets into pseudo 3D training data to expand detection vocabulary. 

However, most of these approaches suffer from limited generalizability, and even advanced models~\cite{brazil2023omni3d,li2024unified3dobjectdetection} are constrained by a closed vocabulary, restricting their ability to recognize or detect object classes that were not included during training. To address this limitation, our work focuses on exploring the potential of monocular open-vocabulary 3D detection.

Concurrent to our work,  DetAny3D~\cite{zhang2025detect} develops a promptable 3D detection model using large-scale data and foundation models. This work addresses similar problems to ours through different methodologies. We compare with DetAny3D in our experiments quantitatively with extensive experiments.

\section{Methodology}
\label{sec:method}

\begin{figure}[t!]
    \centering
    \includegraphics[width=\linewidth]{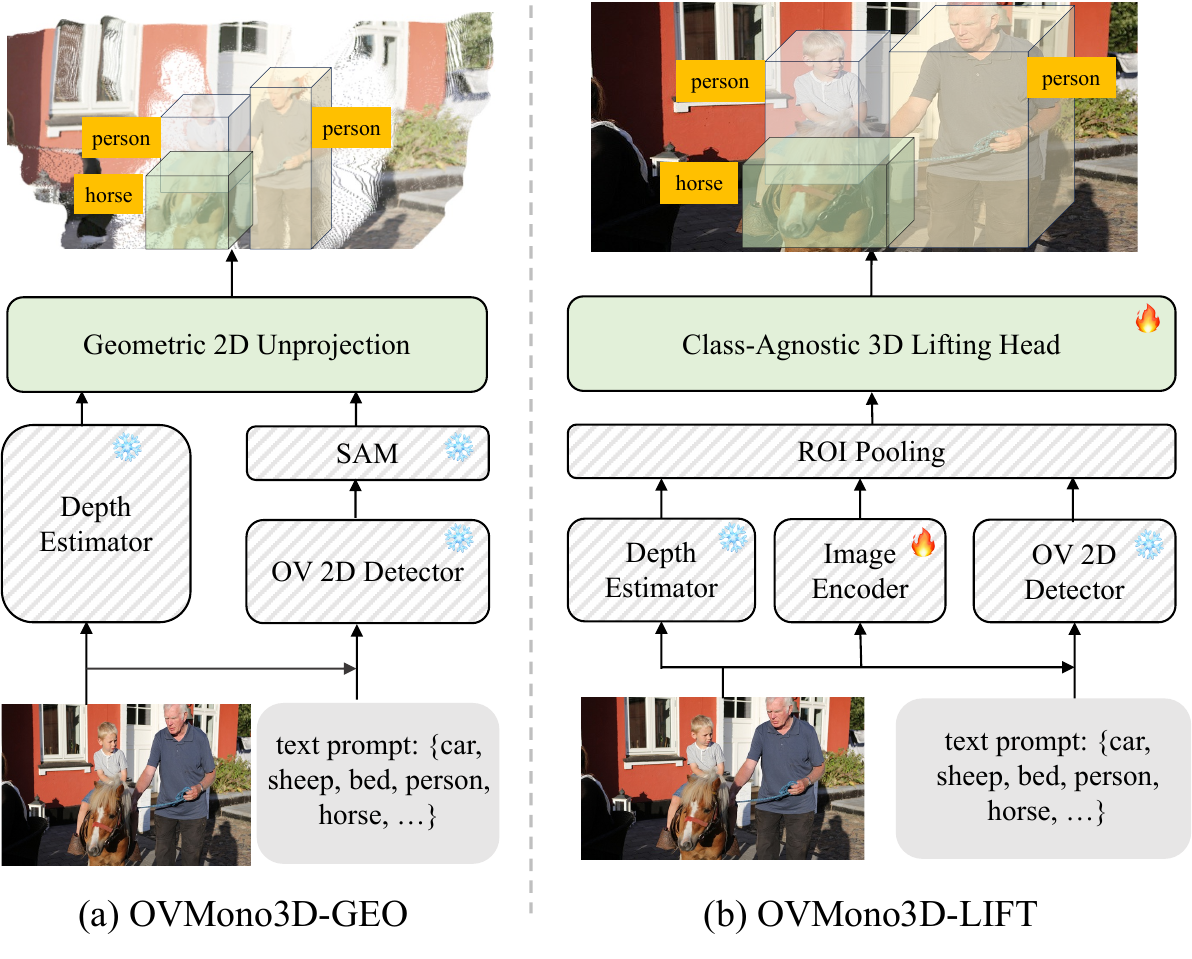}
    % \fbox{\rule{0pt}{150pt} \rule{210pt}{0pt}}
    % \vspace{-3mm}
     \caption{\textbf{Proposed Methods.} \textbf{(a)} \method{}-GEO is a training-free method that predicts 3D detections from 2D via geometric unprojection using off-the-shelf depth estimation (\ie UniDepthv2~\cite{piccinelli2025unidepthv2}), segmentation (\ie SAM~\cite{kirillov2023segment}), and OV 2D detector~\cite{liu2023grounding}.
     \textbf{(b)}~\method{}-LIFT is a learning-based approach that trains a class-agnostic neural network to lift 2D detections and geometric information to 3D. 
     Both approaches decouple 2D recognition and localization from 3D bounding box estimation.
     }
     % \vspace{-3mm}
    \label{fig:method}
\end{figure}

Our approach builds on two frameworks: Cube R-CNN~\cite{brazil2023omni3d} for closed-vocabulary monocular 3D detection and Grounding DINO~\cite{liu2023grounding} for open-vocabulary (OV) 2D detection. In this section, we first provide an overview of these frameworks (Sec.~\ref{subsec:prelimi}); we then introduce our proposed methods for addressing \method{} (Sec.~\ref{subsec:unprojection} -~\ref{subsec:lifting}, ~\cref{fig:method}).

\subsection{Preliminaries}
\label{subsec:prelimi}

\noindent
\textbf{Cube R-CNN}~\cite{brazil2023omni3d} is a state-of-the-art monocular 3D detection model trained on a large-scale 3D dataset (\ie, Omni3D).
It extends Faster R-CNN~\cite{ren2016faster} with a 3D cube head.
Using 2D region proposals as input, the cube head employs ROI poolers to extract local image features and then predicts 3D bounding boxes with MLPs.
The training objective of Cube R-CNN is defined as:
\begin{equation}
\mathcal{L} = \mathcal{L}_{\text{2D}} + \sqrt{2}\, \exp(-\mu)\, \mathcal{L}_{\text{3D}} + \mu,
\label{eq:loss}
\end{equation}
where $\mathcal{L}_{\text{2D}}$ includes the classification and bounding box regression losses from the 2D detection head~\cite{ren2016faster}, $\mathcal{L}_{\text{3D}}$ is the loss from the 3D cube head, and $\mu$ denotes the uncertainty score.
The 3D loss $\mathcal{L}_{\text{3D}}$ consists of disentangled losses for each 3D attribute~\cite{simonelli2019disentangling}:
\begin{equation}
\mathcal{L}_{\text{3D}} = \sum_{a} \mathcal{L}_{\text{3D}}^{(a)} + \mathcal{L}_{\text{3D}}^{\text{all}},
\end{equation}
where $a \in \{(x_\text{2D}, y_\text{2D}),\ z,\ (w, h, l),\ r\}$ represents variable groups for the 2D center shift, depth, 3D dimensions (\ie, width, height, and length in meters), and 3D poses. 
$\mathcal{L}_{\text{3D}}^{(a)}$ isolates the error of a specific group by substituting other predicted variables with ground-truth values when constructing the predicted 3D bounding box $B_{\text{3D}}$, while $\mathcal{L}_{\text{3D}}^{\text{all}}$ compares the vertices of predicted 3D bounding box with the ground truth using Chamfer distance.

\noindent
\textbf{Grounding DINO}~\cite{liu2023grounding} is a leading framework for OV 2D detection, which combines the Transformer-based DINO detector~\cite{zhang2022dino} with grounded pretraining. 
We adopt a pretrained Grounding DINO as our default OV 2D detection model due to its superior performance and strong zero-shot generalization capabilities. 
An ablation study of different base 2D detectors on our approach is provided in Sec.~\ref{sec:exp}.

\vspace{1mm}
\subsection{Training-free \method{}-GEO}
\label{subsec:unprojection}
\vspace{1mm}
We first establish a training-free baseline that unprojects 2D detections into 3D using geometric principles (\cref{fig:method}a).
% We aim at exploring the feasibility of decoupling 2D recognition from 3D estimation without any 3D supervision.
Given an input image $I$, a text prompt $T$, and 2D bounding boxes with category labels $\{ (b_i,\, c_i) \}_{i=1}^N$ predicted by an OV 2D detector~\cite{liu2023grounding}, the method proceeds as follows.

\noindent
\textbf{3D Geometric Prediction.} For each detected object $(b_i,\, c_i)$, an instance segmentation mask $S_i$ is obtained using a segmentation model (\eg, SAM~\cite{kirillov2023segment}), and a metric depth estimation model (\eg, UniDepthv2~\cite{piccinelli2025unidepthv2}) generates the depth map $D \in \mathbb{R}^{H \times W}$. Pixels within $S_i$ are unprojected into 3D space using the camera intrinsic matrix $K \in \mathbb{R}^{3 \times 3}$ to form a point cloud $P_i$. Here, $u$ and $v$ denote the pixel coordinates in the image, and each 3D point $p = [x, y, z]^\top \in P_i$ is computed as:
\[
p = D(u, v) \cdot K^{-1} \begin{bmatrix} u & v & 1 \end{bmatrix}^T.
\]

\noindent
\textbf{3D Bounding Box Generation.} 
To estimate the 3D bounding box parameters $B_i = (t_i,\, d_i,\, r_i)$ from the point cloud  $P_i$, 
we first apply Principal Component Analysis (PCA) on $P_i$ to
estimate the object orientation $r_i$ from the principal component direction. 
PCA determines the object's main orientation by identifying the directions of maximum variance, enabling tight bounding box fitting. 
DBSCAN clustering~\cite{ester1996density} is then applied to remove outliers from noisy depth predictions and imperfect segmentation masks, ensuring robust parameter estimation.  Finally, the centroid $t_i$ and  dimensions $d_i$ are computed from the cleaned point cloud along the estimated orientation $r_i$.

While this method does not require 3D supervision, its accuracy relies heavily on the quality of depth estimation and segmentation, and it is particularly sensitive to occlusions, where incomplete point clouds often lead to inaccurate bounding box reconstruction (see~\cref{fig:geo_sup} in the supplementary).
This approach also shares similarities with the 3D bounding box generation pipeline in OVM3D-Det~\cite{huang2024training}. However, their work employs the generated 3D boxes as pseudo labels to expand the vocabulary of training data, without directly evaluating the accuracy of the 3D box generation itself as a \method{} method.

\vspace{1mm}
\subsection{Data-driven \method{}-LIFT}
\vspace{1mm}

\label{subsec:lifting}

\begin{table*}[!tp]
\centering
\setlength{\tabcolsep}{6pt}
\resizebox{0.99\textwidth}{!}{
\begin{tabular}{@{}
  l
  ccccccc}
\toprule
                  \multirow{2}{*}{Methods}&  \multicolumn{6}{c}{Novel Categories in Omni3D}      & \multicolumn{1}{c}{Novel Dataset }   \\ \cmidrule(lr){2-7} \cmidrule(lr){8-8}
                  & AP$_\threed$\textuparrow & AP$_\threed^\text{15}$\textuparrow & AP$_\threed^\text{25}$\textuparrow & AP$_\threed^\text{50}$\textuparrow & AP$_\threed^\text{easy}$\textuparrow & AP$_\threed^\text{hard}$\textuparrow &  AP$_\threed^\text{city}$ \textuparrow 
                 \\

\midrule
% \emph{Geometric 2D unprojection:}    &               &               &               &               &               &               \\
% G-DINO-3D (FT) &6.77& 4.43         & 6.21         & 4.92           &  1.12  & 4.77 & 4.24        \\
% G-DINO-3D (FZ)          &19.22              &3.54           & 5.74            &2.71              & 0.07   & 4.59 & 2.96              \\
% 3D GroundDINO (FT)    &6.23 &3.16  &4.61  &3.08  &0.58  &2.90&3.30 \\
% \midrule
% \emph{Class-agnostic 3D lifting:} &               &               &               &               &               &               \\

% \midrule
% \emph{Extending OV 2D detector:}    &               &               &               &               &               &               \\
OVM3D-Det~\cite{huang2024training} & 6.74 / 18.30 & 11.35 / 29.69 &  5.95 / 15.87 &  0.73 / 2.31 & 9.70  / 22.99 &  5.05 / 15.63 &  0.04 / 0.05   \\
DetAny3D~\cite{zhang2025detect} & 9.22 / 21.44 & - & - & - & - & - & 11.05 / 15.71  \\
\method{}-GEO &  8.48 / 20.63   &  14.07 / 35.26  & 8.22 / 18.26   &  0.49 / 1.01  & 9.71 / 21.20  & 7.78 / 20.30 &  0.73  / 0.90  \\
% \method{}-LIFT (freeze dino)    & 16.04 &24.57 &14.99 &2.21 &21.05  & 13.18 \\
\method{}-LIFT* & 7.31 / 18.61 & 11.47 / 27.95 & 7.06  / 17.62 & 0.93 / 3.48 & 11.39 / 26.14 & 4.97 / 14.31  & 6.57 / 10.21  \\  
% \method{}-FUSION (text thres 0.25) &  7.94/ 24.08 & / 37.45 & / 23.24 & / 4.00 & / 30.13 & / 20.62    \\
% \method{}-FUSION (text thres 0.01) &  9.61/ 24.08 & 15.00 / 37.45 & 9.39 / 23.24 & 1.41 / 4.00 &  14.71 / 30.13 &  6.7 / 20.62    \\
\method{}-LIFT &  \textbf{9.61} / \textbf{24.08} & \textbf{15.00} /\textbf{ 37.45} & \textbf{9.39} / \textbf{23.24} & \textbf{1.41} / \textbf{4.00} &  \textbf{14.71} / \textbf{30.13} &  \textbf{6.7} / \textbf{20.62}  & \textbf{12.88} / \textbf{16.76}    \\
\bottomrule
\end{tabular}
}
\caption{\textbf{Performance on Novel Categories and Datasets.} We report  AP$_\threed$ on Omni3D novel categories and novel datasets, including results under IoU${_\text{3D}}$  thresholds of 0.15, 0.25, and 0.5, and across easy and hard categories splits for novel categories in Omni3D (\cref{subsec:dataset}
). All metrics are reported in the format of standard metric / target-aware metric (\cref{subsec:evaluation_prototype}). We notice that with UniDepthv2~\cite{piccinelli2025unidepthv2}, performance drops significantly on CityScapes3D~\cite{gahlert2020cityscapes}; therefore, we use Metric3Dv2~\cite{hu2024metric3d} for all methods requiring pseudo depth input on this dataset only.
\textbf{Bold} indicates best results and ``-'' denotes unreported results in previous works. }
\label{tab:main}
\end{table*}

% here we can only report the cubercnn 2d input results. in supplementary we can report grounding dino input, otherwise it will be hard to compare with previous baselines...

% commented out the following table because it is under gdino input and target-aware metric

\iffalse

\begin{table*}[!tp]
\centering
% \setlength{\tabcolsep}{10pt}
\resizebox{0.6\linewidth}{!}{
\begin{tabular}{c|cccccc|ccc}
\toprule
      Method           & AP$_\threed^\text{sun}$& AP$_\threed^\text{hyp}$&AP$_\threed^\text{obj}$&AP$_\threed^\text{ark}$& AP$_\threed^\text{kit}$&AP$_\threed^\text{nus}$& AP$_\threed^\text{all}$ \\ 
\midrule
SMOKE \\
FCOS3D \\
ImVoxelNet \\
Cube R-CNN~\cite{brazil2023omni3d}         & 21.36  & 8.66  & 54.38  & 49.69 &  37.25 & \textbf{31.95} & 26.78  \\ 
OVM3D-Det* &19.82 & 8.97 & 11.50 &33.38 & 13.30 &10.45 &12.35 \\
\method{}-GEO & 9.30 & 8.07 & 5.13&1.34 & 2.85&3.01&  4.03\\ 
% \method{}-LIFT & 24.18 & 12.19 & 57.69 &45.65&28.22&22.09& 24.77    \\ 
\method{}-LIFT & 27.61 & 10.44 & 62.01 & 48.66 & 29.06 & 24.89 & 26.39         \\
\method{}-FUSION & \textbf{31.86} &\textbf{12.91} & \textbf{62.58} & \textbf{53.06} & \textbf{39.21} & 27.21 & \textbf{29.28} \\
\bottomrule
\end{tabular}
}
% \vspace{-6pt}
\caption{\textbf{Performance on Base Categories.} We report the AP$_\threed$ on six subsets of Omni3D and the overall score, using target-aware metric. 
}
\label{tab:closed_vocab_result}
\end{table*}

\fi

\begin{table*}[!tp]
\centering
\resizebox{0.99\linewidth}{!}{
\begin{tabular}{l|rrrrrr|rrrrr|r}
\toprule
      Method         & AP$_\threed^\text{kit}$\textuparrow &AP$_\threed^\text{nus}$\textuparrow  & AP$_\threed^\text{sun}$\textuparrow & AP$_\threed^\text{hyp}$\textuparrow  &AP$_\threed^\text{ark}$\textuparrow &AP$_\threed^\text{obj}$\textuparrow & AP$_\threed^\text{25}$\textuparrow & AP$_\threed^\text{50}$\textuparrow & AP$_\threed^\text{near}$\textuparrow & AP$_\threed^\text{med}$\textuparrow & AP$_\threed^\text{far}$\textuparrow &  AP$_\threed^\text{all}$\textuparrow \\ 
\midrule
\textbf{Closed-vocabulary methods} \\
% M3D-RPN & 10.4 & 17.9 \\ 
% SMOKE & 25.4 & 20.4 \\
\ \ \  SMOKE~\cite{liu2020smoke} & - &  - & - & - & - & - & - & - & - & - & - & 9.60 \\
% FCOS3D & 14.6 & 20.9 &  \\
\ \ \  FCOS3D~\cite{wang2021fcos3dfullyconvolutionalonestage}  & - &  - & - & - & - & - & - & - & - & - & - & 9.80  \\
% PGD & 21.4 & 26.3 & \\
\ \ \  PGD~\cite{wang2022probabilistic} & - &  - & - & - & - & - & - & - & - & - & - & 11.20 \\
% GUPNet & 24.5 & 20.5 & \\
% ImVoxelNet & 23.5 & 23.4 \\
\ \ \  ImVoxelNet~\cite{rukhovich2022imvoxelnet} & - &  - & - & - & - & - & - & - & - & - & - & 9.40  \\
% BEVFormer & 23.9 & 29.6 & \\
% PETR & 30.2 & 30.1 & \\
\ \ \   Cube R-CNN~\cite{brazil2023omni3d} &\textbf{32.47} &29.98 & 15.23 &7.39 & 41.72 & 53.85 & 25.51 & 9.59 & 28.32 & 12.05 & 8.50 &  23.68  \\ 
\ \ \   UniMODE~\cite{li2024unified3dobjectdetection} & 29.20 & \textbf{36.00} & \textbf{23.00} & \textbf{8.10} & \textbf{48.00} & \textbf{66.10} & \textbf{30.20} & \textbf{10.60} & \textbf{31.10} & \textbf{14.90} & \textbf{8.70} & \textbf{28.20}     \\
\midrule
\textbf{Open-vocabulary methods} \\
\ \ \   OVM3D-Det~\cite{huang2024training} & 7.59 & 14.25 & 11.82 & 5.34 & 25.23 & 1.93 & 9.33 & 1.36 & 10.51 & 4.82 & 3.71 & 9.49 \\
\ \ \  DetAny3D~\cite{zhang2025detect} & \textbf{31.61} & \underline{30.97} & 18.96 & 7.17 &46.13 & 54.42 & - & - & - & - & - & 24.92  \\

\ \ \  \method{}-GEO & 1.22 & 3.23 & 9.99 & 3.98 & 21.03 & 9.22 & 5.98  & 0.23 & 9.94 & 1.59 & 0.15 & 6.64 \\ 
% \method{}-LIFT & 24.18 & 12.19 & 57.69 &45.65&28.22&22.09& 24.77    \\ 
% \method{}-LIFT (cuber 2d input) & 26.98 & 27.12 & 17.31 & 7.80 & 45.00 & 63.44 & 27.02 &11.56 & 30.47 & 10.73 & 7.04 & 25.15       \\
\ \ \  \method{}-LIFT* & 25.58 & 30.56 & \underline{19.52} & \underline{9.92} & \underline{49.34} & \underline{62.72} & \underline{29.04} & \underline{11.91} & \underline{33.96} & \underline{12.20} & \underline{7.21} & \underline{27.09}    \\
% \method{}-FUSION (cuber 2d input) & 33.77 & 28.50 & 19.12 &8.98 & 49.08 & 64.25 & 29.40 & 12.93 & 32.04 &  11.79 & 8.82 & 27.13 \\
\ \ \  \method{}-LIFT & \underline{31.44} & \textbf{32.47} & \textbf{23.24} & \textbf{11.89} & \textbf{54.21} & \textbf{63.48} & \textbf{32.23} & \textbf{12.90} & \textbf{37.51} & \textbf{13.35} & \textbf{8.96} & \textbf{29.63} \\
\bottomrule
\end{tabular}
}
\vspace{2pt}
\caption{\textbf{Performance on Base Categories.} We report the AP$_\threed$ on six subsets of the Omni3D test set, including results under IoU${_\text{3D}}$  thresholds of 0.25 and 0.5, and across different distances (near, medium, far). AP$_\threed^{\text{all}}$ denotes the overall score averaged across all subsets. For open-vocabulary methods without 2D heads, we use 2D box predictions from Cube R-CNN as input. All the metrics are in standard evaluation. \textbf{Bold} indicates best results, \underline{underlined} indicates second best, and ``-'' denotes unreported results in previous works.}

\label{tab:closed_vocab_result}
\end{table*}

% Specifically, we introduce 
To overcome the limitations of the geometric-based approach, we propose \method{}-LIFT (\cref{fig:method}b), which learns to estimate 3D bounding boxes from large-scale 3D annotations. This framework decouples \method{} into two stages:
(1) recognizing and localizing objects in 2D with off-the-shelf open-vocabulary detectors, and
(2) class-agnostically lifting the 2D bounding boxes into 3D cuboids.
We detail the key components below.

\noindent
\textbf{Image Encoder.} \method{}-LIFT employs a pretrained DINOv2~\cite{oquab2023dinov2} to extract image features, and
% $F \in \mathbb{R}^{\frac{H}{14} \times \frac{W}{14} \times C}$, where $C$ is the feature dimension. 
a Feature Pyramid module~\cite{li2022exploring} to generates multi-scale hierarchical feature pyramid. 
The predicted 2D bounding boxes are used as regions of interest for ROI pooling, extracting local object features of size $7 \times 7$. 
These features are fed into a cube head, as described in~\cref{subsec:prelimi}, to predict 3D attributes.

\noindent
\textbf{Depth Estimator.} Similar to \method{}-GEO, the method first obtain the predicted depth map $D \in \mathbb{R}^{H \times W}$, then unproject it to a point map $P \in \mathbb{R}^{3 \times H \times W}$ using the camera intrinsic matrix $K \in \mathbb{R}^{3 \times 3}$. ROI pooling is then performed on the point map using the 2D bounding boxes to extract local geometric information of size $ 3 \times 7 \times 7$. This geometric information is concatenated with the visual ROI features to form geometry-informed features, which are input into the 3D cube head to predict 3D box attributes.
During both training and inference, the method uses pseudo depth maps obtained from a pretrained metric depth estimator—no ground truth depth maps are required. 
By fusing geometric information into the model, the network can better infer object depth and dimensions, producing more accurate predictions.

\noindent
\textbf{Class-agnostic 3D Lifting head.} Unlike Cube R-CNN~\cite{brazil2023omni3d}, our 3D cube prediction head is \emph{class-agnostic}, while their method employs class-specific layers and per-class average size statistics, which limit generalization to novel categories. 
% \todo{The prediction head is simply implemented as ...}  
The model trains only on base categories with the training objective identical to Cube R-CNN, as described in~\cref{subsec:prelimi}. Since the method also train a 2D head on base categories, the model preserves detection ability on these categories. During inference, for novel categories, an OV 2D detector (\eg, Grounding DINO~\cite{liu2023grounding}) is used to obtain 2D boxes, which serve as input to the ROI pooling module. 

\paragraph{\method{}-LIFT*.} We include a model variant as a baseline. It removes the depth estimation module from the full \method{}-LIFT framework, relying solely on visual features from the image encoder and Feature Pyramid Network for 3D bounding box prediction. This variant serves to evaluate the contribution of geometric depth information by comparing performance against the full model that incorporates both visual and geometric features.

\section{Evaluation Metrics}
\label{subsec:evaluation_prototype}

\begin{figure}[!t]
    \centering
    \includegraphics[width=0.97\columnwidth]{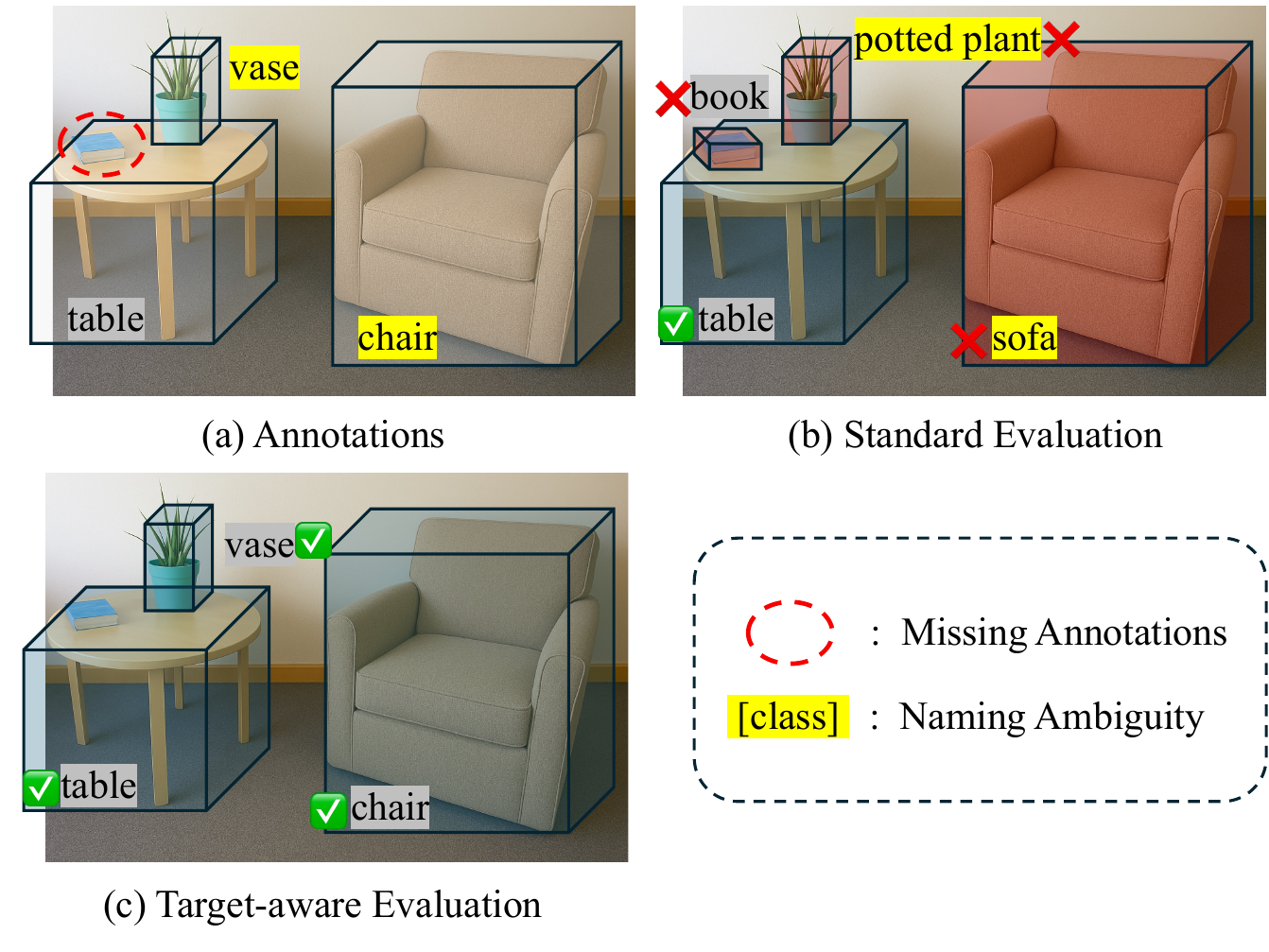}
    % \vspace{-3mm}
    \caption{By prompting only categories that exist in the annotations, our target-aware evaluation mitigates the negative impact of missing annotations  (\eg, ``book" in (a) ) and naming ambiguity (\eg, ``vase" \textit{vs.}  ``potted plant'' and ``chair'' \textit{vs.}  ``sofa''.)
    }
    \label{fig:target_aware_examples}
\end{figure}

\paragraph{Standard Metrics.}
Mean Average Precision (AP$_\threed$) based on Intersection over Union (IoU) is the widely used standard metric for closed-vocabulary 3D object detection tasks~\cite{brazil2023omni3d, li2024unified3dobjectdetection}. However, directly applying traditional evaluation protocols to \method{} presents challenges due to common issues in 3D dataset annotations:

\noindent
\textit{Missing annotations:} Comprehensive 3D annotation is often impractical due to high costs. Fig.~\ref{fig:target_aware_examples}a illustrates an example of this issue, where the book is not labeled.

\noindent
\textit{Naming ambiguity:} Objects may be labeled with different naming conventions or annotation policies during annotation (\eg, table \textit{vs.} desk). Standard open-vocabulary 2D detection methods typically prompt 2D detectors with exhaustive lists of possible categories, which can lead to correct predictions with class names that do not align with the dataset annotations, especially for vaguely defined or overlapping categories, as shown in~\cref{fig:target_aware_examples} (chair \textit{vs.}  sofa, vase \textit{vs.}  potted plant).

\paragraph{Target-Aware Metrics.}  
To address these issues, we propose a simple evaluation approach that considers only the categories with ground-truth labeled instances in each image. Specifically, instead of providing the 2D detector with an exhaustive list of possible categories, we only prompt it with category names that exist in the annotations for each image. Since human annotators typically label all instances of the same category within a single image, categories with ground-truth annotations are likely to be fully annotated.
\cref{fig:target_aware_examples} illustrates our approach enables more accurate model assessment despite annotation issues. For instance, the example highlights missing annotations for  ``books," and naming ambiguities between ``chair" and ``sofa." This validates the practicability of our evaluation under annotation issues.

\begin{figure*}[ht!]
    \centering
    \includegraphics[width=1.0\textwidth]{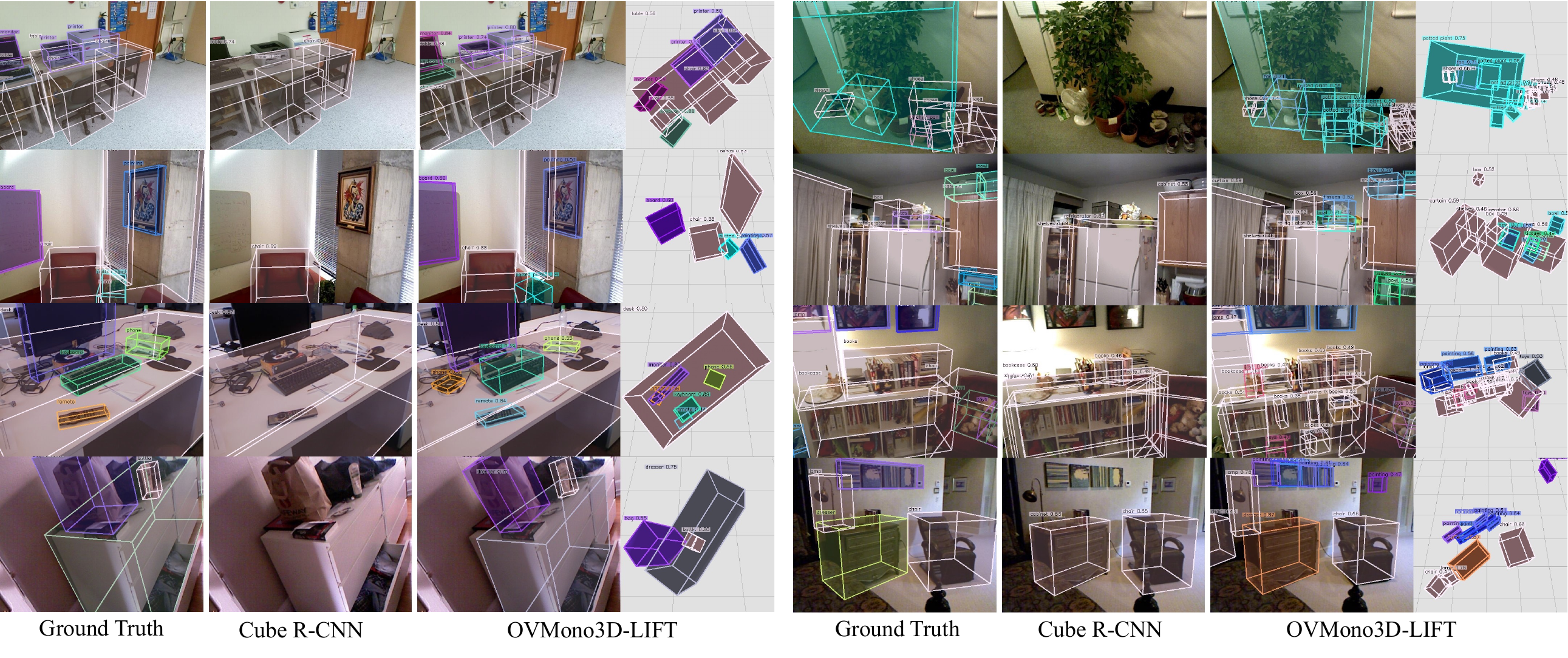}
    \vspace{-5mm}
    \caption{\textbf{Qualitative Visualizations on the Omni3D Test Set.} For each example, we present the ground truth annotations, the predictions of Cube R-CNN and \method{}-LIFT, displaying the 3D predictions overlaid on the image. For \method{}-LIFT, we also present a top-down view with a base grid of $1\,\text{m} \times 1\,\text{m}$ tiles. Base categories are depicted with \textcolor{brown}{brown} cubes, while novel categories are represented in other colors. Zoom in for best viewing. See~\cref{sec:more_qualitative} for more visualizations. 
    }
    \vspace{-3mm}
    \label{fig:qualitative}
\end{figure*}
\section{Experiments}
\label{sec:exp}

\subsection{Experimental Setup}%
\label{sec: implementation}

\paragraph{Datasets.} 
\label{subsec:dataset}
The experiments are conducted using the Omni3D~\cite{brazil2023omni3d} dataset, the largest dataset for monocular 3D object detection across both indoor and outdoor scenes. Omni3D is repurposed and combined from six established sources including SUN RGB-D~\cite{song2015sun}, ARKitScenes~\cite{baruch2021arkitscenes}, Objectron~\cite{ahmadyan2021objectron}, Hypersim~\cite{roberts2021hypersim}, nuScenes~\cite{caesar2020nuscenes}, and KITTI~\cite{geiger2012we}. 
It comprises a substantial 234k images, 3 million labeled 3D bounding boxes, and covers 98 distinct object categories. 

During training, we use the 50 categories Cube R-CNN~\cite{brazil2023omni3d} trained on as the base categories. 
For evaluation, we select 22 categories from the remaining classes as novel categories. 
These categories are chosen based on two criteria: the number of test instances and the precision of category naming. To facilitate a detailed assessment of zero-shot generalization capabilities, we further divide these categories into \emph{easy} and \emph{hard} subsets according to object visibility. See~\cref{sec:per_cat} for detailed splits. 

Besides evaluating novel categories in Omni3D~\cite{brazil2023omni3d}, we perform zero-shot evaluation on a novel dataset with novel camera models. CityScapes3D~\cite{gahlert2020cityscapes} includes base categories from the self-driving domain. We use the same evaluation split as DetAny3D~\cite{zhang2025detect}.

\vspace{-5pt}\paragraph{Baselines.} For novel categories, we compare against OVM3D-Det~\cite{huang2024training}, which generates pseudo 3D labels by combining Grounding DINO~\cite{liu2023grounding}, SAM~\cite{kirillov2023segment}, UniDepth~\cite{piccinelli2024unidepth}, and LLM-generated size priors. We evaluate their labeling pipeline as a baseline detector against ground truth annotations. We also include comparisons with DetAny3D~\cite{zhang2025detect}.
For base categories, we compare against closed-vocabulary methods including SMOKE~\cite{liu2020smoke}, FCOS3D~\cite{wang2021fcos3dfullyconvolutionalonestage}, PGD~\cite{wang2022probabilistic}, ImVoxelNet~\cite{rukhovich2022imvoxelnet}, Cube R-CNN~\cite{brazil2023omni3d}, and UniMODE~\cite{li2024unified3dobjectdetection}, as well as open-vocabulary methods including OVM3D-Det~\cite{huang2024training} and DetAny3D~\cite{zhang2025detect}.

\vspace{-5pt}\paragraph{Evaluation Metrics.}

We report the mean Average Precision in 3D (AP$_\threed$) computed across all evaluated categories. Following the Omni3D~\cite{brazil2023omni3d} evaluation protocol, predictions are matched to ground truth using 3D Intersection over Union (IoU${_\text{3D}}$), which measures the volumetric overlap between predicted and ground truth 3D bounding boxes. The evaluation is conducted across IoU${_\text{3D}}$ thresholds $\tau \in [0.05,0.10,...,0.50]$, with the final AP$_\threed$ representing the mean average precision across all thresholds and categories. For evaluation on novel categories and novel datasets, we additionally report target-aware metrics~(see~\cref{subsec:evaluation_prototype}).

\vspace{-5pt}\paragraph{Implementation Details.} 
Our implementation is based on PyTorch3D~\cite{ravi2020accelerating} and Detectron2~\cite{wu2019detectron2}. We initialize the image feature encoder with pre-trained DINOv2 base weights~\cite{oquab2023dinov2}. The model is trained on eight NVIDIA A100 GPUs for five days. For reference, DetAny3D~\cite{zhang2025detect} is trained on 64 A100 GPUs for two weeks.
The models are trained for 116k steps with a batch size of 64 using SGD optimizer with an initial learning rate of 0.0012, which decays by a factor of 10 at 60\% and 80\% of training. We apply standard image augmentations including random horizontal flipping and resizing during training. 

Due to computational constraints, models trained in the analysis section (~\cref{sec:analysis}) employ a resource-efficient setting where the pre-trained image encoder is kept frozen during training and the depth estimator is excluded.
Unless explicitly stated otherwise, all methods use Grounding DINO~\cite{liu2023grounding} for 2D detection on novel categories, and all methods requiring pseudo depth input use UniDepthv2~\cite{piccinelli2025unidepthv2}.

\subsection{Model Performance}

Fig.~\ref{fig:qualitative} shows qualitative results on the Omni3D test set. 
In comparison with Cube R-CNN~\cite{brazil2023omni3d}, \method{}-LIFT detects not only objects of base categories, but also novel categories that are unseen at the training time. 

\vspace{-5pt}\paragraph{Novel Category and Dataset Performance.}
 \cref{tab:main} shows that \method{}-LIFT achieves state-of-the-art performance on novel categories and novel datasets.

Geometry-based methods OVM3D-Det~\cite{huang2024training} and \method{}-GEO show lower performance, particularly on Cityscapes3D~\cite{gahlert2020cityscapes}, due to their sensitivity to occlusions and noisy depth estimation.
The performance improvement from LIFT* to LIFT underscores the importance of incorporating additional geometric information from foundation depth estimators into the model. This enhanced performance demonstrates that our geometry-informed 2D lifting design effectively exploits 2D data-driven priors, including OV 2D detectors, self-supervised features, and metric depth estimators. By decomposing the task into OV 2D detection and 2D-to-3D lifting, our approach addresses challenges associated with low-quality and limited 3D annotations.

\vspace{-5pt}\paragraph{Base Category Performance.}

\cref{tab:closed_vocab_result} compares \method{}-LIFT with baselines on base categories. \method{}-LIFT achieves the best performance among both closed-vocabulary and open-vocabulary methods, with an overall  AP$_\threed$ of 29.63. It surpasses Cube R-CNN~\cite{brazil2023omni3d} by 5.95 points and UniMODE~\cite{li2024unified3dobjectdetection} by 1.43 points, while additionally providing generalization to novel categories. Compared to the open-vocabulary baseline DetAny3D~\cite{zhang2025detect}, it achieves a 4.71-point improvement, demonstrating strong performance on base categories. Notably, \method{}-LIFT* also achieves competitive results with 27.09 overall AP$_\threed$, ranking second among open-vocabulary methods. These results validate the effectiveness of our approach.

\subsection{Analysis}
\label{sec:analysis}

\begin{table}[t!]
\centering
\resizebox{0.99\columnwidth}{!}{
\begin{tabular}{@{}llcccc@{}}
\toprule
  Methods &  IoU${_\text{3D}}$\textuparrow & XY\textuparrow & Depth\textuparrow & Size\textuparrow & Pose\textuparrow \\ 
\midrule
 \method{}-GEO &  18.69 & 47.96  & 30.97  & 18.68  & 37.84  \\
 \method{}-LIFT* & 18.11  & 49.39 & 27.23  & 48.10 & 72.44  \\
 \method{}-LIFT & \textbf{22.03} & \textbf{52.91} & \textbf{32.59} & \textbf{49.64} & \textbf{73.57} \\
\bottomrule
\end{tabular}
% \vspace{-2mm}
}
\caption{\textbf{Disentangled Metrics on Novel Categories.} We report the overall and disentangled IoU${_\text{3D}}$ (\%) for different attribute predictions on Omni3D's novel categories. 
% Since NHD is only computed on high-IoU predictions --- which usually vary across different models ---- comparisons are valid only within each model. Lower values indicate better performance.
For fair comparison, same 2D ground truth box inputs are applied for all methods.
}
\vspace{-4mm}
\label{tab:disentangle}
\end{table}

\paragraph{Disentangled Metrics.} 
To analyze the contributions of different predicted attributes to 3D bounding box errors, we report disentangled IoU${_\text{3D}}$ for position (XY), depth, dimensions, and pose. For each attribute, we compute the IoU${_\text{3D}}$ of a modified 3D bounding box where only that attribute is predicted, while all others are set to their ground-truth values. The IoU${_\text{3D}}$ is then computed against the ground-truth cube. To facilitate the comparison, for all methods we use ground truth 2D box as input. 

\cref{tab:disentangle} exhibits the overall and disentangled IoU${_\text{3D}}$ for different attribute predictions. For LIFT* and LIFT methods, object depth prediction consistently contributes the most error to the overall prediction. This indicates that object depth estimation is the primary bottleneck of our task. Furthermore, the size and pose predictions in GEO method exhibit larger errors, indicating that geometry-based methods are not effective for these predictions. This underscores the necessity of developing learning-based methods.

\vspace{-5pt}\paragraph{2D Bounding Box Input.} 
\cref{tab:ablation_2d} evaluates the impact of 2D detectors on both 2D and 3D detection  performance for novel categories.
We test two state-of-the-art 2D detectors: YOLO-World~\cite{cheng2024yolo} and Grounding DINO~\cite{liu2023grounding}. 
Grounding DINO consistently outperforms YOLO-World across all evaluation metrics,
demonstrating that Grounding DINO is a preferable off-the-shelf 2D detector for \method{}.

\begin{table}[!tp]
\centering
\resizebox{0.99\columnwidth}{!}{
\begin{tabular}{@{}
  l
  ccccc}
\toprule
                  2D Box Input& AP$_\twod$\textuparrow & AP$_\threed$\textuparrow & AP$_\threed^\text{15}$\textuparrow & AP$_\threed^\text{25}$\textuparrow & AP$_\threed^\text{50}$\textuparrow \\

\midrule
% \emph{Class-agnostic 3D lifting:} &               &               &               &               &               &               \\
 % \textcolor{gray}{\emph{Ground Truth}} & \textcolor{gray}{100} & \textcolor{gray}{17.21} & \textcolor{gray}{25.71} & \textcolor{gray}{16.30} & \textcolor{gray}{2.79} \\ 
 % \arrayrulecolor{gray} \midrule
 % \midrule
YOLO-World~\cite{cheng2024yolo} & 19.99 & 20.31 & 31.50 & 19.42 & 2.85
\\
 Grounding DINO~\cite{liu2023grounding}  &\textbf{21.44} &\textbf{24.08} &\textbf{37.45} &\textbf{23.24} &\textbf{4.00}
\\
\bottomrule
\end{tabular}
}
\caption{\textbf{Ablation on 2D Bounding Box Input.} Evaluation of open-vocabulary 2D detectors and their impact on 2D and 3D detection performance for novel categories in Omni3D. }
\vspace{-3mm}
\label{tab:ablation_2d}
\end{table}

\begin{figure}[!t]
    \centering
    \includegraphics[width=0.97\columnwidth]{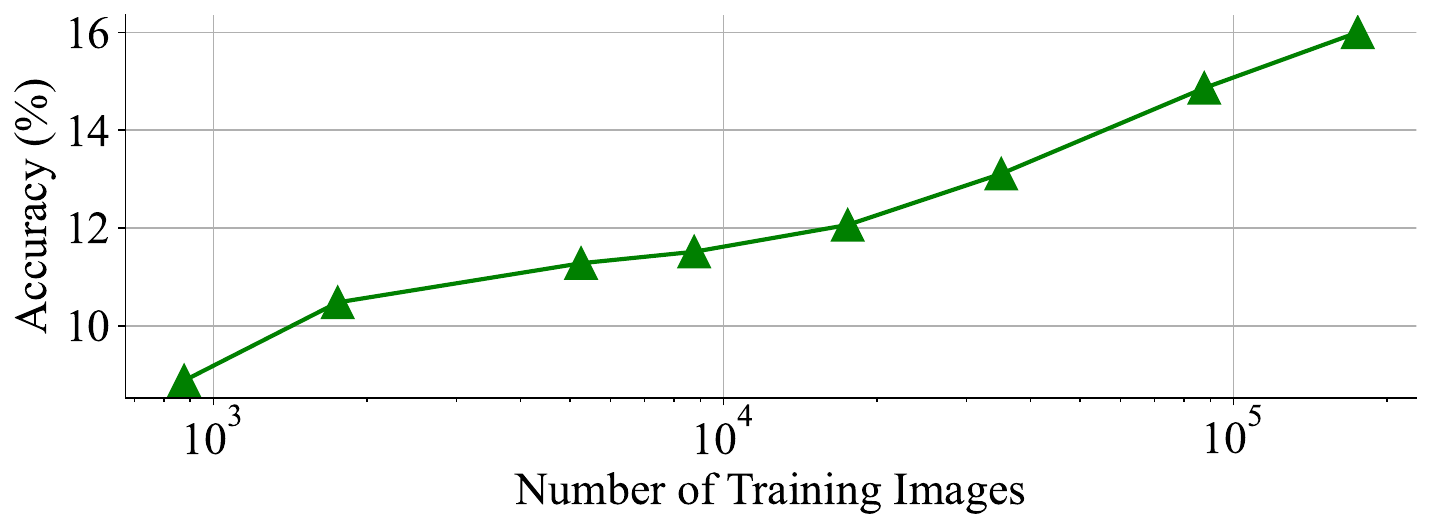}
    \vspace{-2mm}
    \caption{The performance of \method{}-LIFT as a function of training data amount. We report the AP$_\threed$ (\%) on novel categories as the evaluation metric.}
    \label{fig:scaling}
    \vspace{-4mm}
\end{figure}

\vspace{-5pt}\paragraph{Training Data Scaling Law.}
\cref{fig:scaling} reports our model’s  AP$_\threed$ score as a function of the training data size. 
This underscores the critical importance of dataset size in \method{} tasks and suggests that our model may achieve even better performance with more extensive training data.

\paragraph{Pre-trained feature extractor Selection. }

\cref{tab:ablation_feature} shows the impact of pretrained feature extractors on 3D detection performance for novel categories.   DINOv2 achieves the best performance across all evaluation metrics. This result underscores the effectiveness of DINOv2's representations for the 3D detection task. Our findings align with recent studies that highlight DINOv2's strong zero-shot capabilities in understanding depth, multi-view correspondences, and relative pose~\cite{krishnan2025omninocs, goodwin2022zero, zhang2024tale}, indicating that DINOv2's 3D-aware image features are highly suitable for this task.

For a more detailed analysis on the role of synthetic data and quantitative evaluation of naming ambiguity issues in current benchmarks, refer to~\cref{sec:more_analysis}.

\begin{table}[!tp]
\centering
\setlength{\tabcolsep}{6pt}
\resizebox{0.99\columnwidth}{!}{
\begin{tabular}{@{}
  l
  ccccccc}
\toprule
                  Feat. Extractor& Supervision 
                  & AP$_\threed$\textuparrow & AP$_\threed^\text{15}$\textuparrow & AP$_\threed^\text{25}$\textuparrow & AP$_\threed^\text{50}$\textuparrow    \\ 

\midrule
% MOCO-v2 & R50 & SSL & ImageNet1k\\
MAE~\cite{he2022masked}        &  SSL        &7.72&11.95&6.43&0.80                     \\
CLIP~\cite{radford2021learning}& VLM &9.02&14.19&8.44&0.50
\\
MiDas~\cite{ranftl2020towards}       &  Depth        & 10.65&17.42&9.87&0.69
\\
% Cube R-CNN~\cite{brazil2023omni3d}          &  DLA34     & Detection &    Omni3D          & 11.21&15.59&10.13&1.32             \\
DINOv2~\cite{oquab2023dinov2} & SSL &\textbf{16.04} &\textbf{24.57} &\textbf{14.97} &\textbf{2.21}  
\\
\bottomrule
\end{tabular}
}
\caption{\textbf{Ablation Study on Feature Extractors for \method{}-LIFT in 3D Detection on Novel Categories.} We report AP$_\threed$ for 
 various feature extractors with frozen parameters during training.}
\label{tab:ablation_feature}
\vspace{-4mm}
\end{table}

\subsection{Zero-Shot Generalization Performance }

\cref{fig:coco} presents \method{}-LIFT prediction on in-the-wild images from COCO~\cite{lin2014microsoft}. 
The results show 2D projections of our predictions are well-aligned with the objects, 
and their top-down views closely match the visual scene layout. 
Even on completely out-of-distribution categories such as elephant and apple, our method produces promising results. This suggests that our model demonstrates zero-shot generalization ability on real-world images. See~\cref{fig:coco_sup} in Supplementary Material for more qualitative results on COCO.

\section{Discussion}
\label{sec:conclusion}

We investigate \method{}, an under-explored task of recognizing and localizing objects from any categories in 3D using a single image. 
We identify unique challenges, notably data scarcity and limitations of standard evaluation metrics.  
We propose simple yet effective approaches, including geometry-based \method{}-GEO and learning-based \method{}-LIFT --- both decouple 2D detection from 3D bounding box prediction.
Such design enables the full utilization of off-the-shelf open-vocabulary 2D detectors pre-trained on large-scale datasets. 
Our analysis pinpoints the key components in the framework,
including 3D-aware image features, base 2D detectors, and the impact of dataset scaling. 
We further demonstrate the zero-shot generalizability of our approach on in-the-wild images. 
Our findings suggest that dataset scale and accurate depth perception remain the major bottlenecks in this task. 
One promising direction may be to develop unsupervised learning that harness the abundance of unlabeled images.
We hope this work inspires future research toward advancing this task.

\section{Limitations}
\label{sec:limitations}
Due to the lack of 3D detection ground truth labels in COCO~\cite{lin2014microsoft}, the qualitative zero-shot evaluation is not feasible to perform. Additionally, our method requires accurate camera intrinsics as input; however, for in-the-wild images, the estimated intrinsics can be inaccurate, leading to errors in prediction. 
Furthermore, the use of computationally heavy components, such as Grounding DINO~\cite{liu2023grounding} and DINOv2~\cite{oquab2023dinov2}, results in slower inference speed compared to Cube R-CNN~\cite{brazil2023omni3d}, which should be addressed in future work. See~\cref{sec:fail} for visualizations of failure cases.

\section{Acknowledgement}

The authors acknowledge the University of Virginia Research Computing and Data Analytics Center, Advanced Micro Devices AI and HPC Cluster Program, Advanced Cyberinfrastructure Coordination Ecosystem: Services \& Support (ACCESS) program, and National Artificial Intelligence Research Resource (NAIRR) Pilot for computational resources, including the Anvil supercomputer (National Science Foundation award OAC 2005632) at Purdue University and the Delta and DeltaAI advanced computing resources (National Science Foundation award OAC 2005572). This work was partly supported by an Adobe Research Gift. 

{
    \small
    \bibliographystyle{ieeenat_fullname}
    \bibliography{main}
}
\clearpage
\appendix
\setcounter{page}{1}
\maketitlesupplementary

\iffalse

\section{Rationale}
\label{sec:rationale}
% 
Having the supplementary compiled together with the main paper means that:
% 
\begin{itemize}
\item The supplementary can back-reference sections of the main paper, for example, we can refer to \cref{sec:intro};
\item The main paper can forward reference sub-sections within the supplementary explicitly (e.g. referring to a particular experiment); 
\item When submitted to arXiv, the supplementary will already included at the end of the paper.
\end{itemize}
% 
To split the supplementary pages from the main paper, you can use \href{https://support.apple.com/en-ca/guide/preview/prvw11793/mac#:~:text=Delete%20a%20page%20from%20a,or%20choose%20Edit%20%3E%20Delete).}{Preview (on macOS)}, \href{https://www.adobe.com/acrobat/how-to/delete-pages-from-pdf.html#:~:text=Choose%20%E2%80%9CTools%E2%80%9D%20%3E%20%E2%80%9COrganize,or%20pages%20from%20the%20file.}{Adobe Acrobat} (on all OSs), as well as \href{https://superuser.com/questions/517986/is-it-possible-to-delete-some-pages-of-a-pdf-document}{command line tools}.

\section{Detailed Algorithm Description for \method{}-GEO}

\fi

Sec.~\ref{sec:per_cat} presents per-category performance on novel classes for \method{}-GEO and \method{}-LIFT. Sec.~\ref{sec:more_qualitative} provides additional qualitative visualizations on various datasets and compares predictions between \method{}-GEO and \method{}-LIFT. Sec.~\ref{sec:fail} discusses failure cases, highlighting challenges such as occlusions, out-of-distribution objects, and small or distant instances. Sec.~\ref{sec:more_analysis} provides more analysis on synthetic data and the naming ambiguity issue in current benchmarks.

\section{Per-category Performance on novel classes}
\label{sec:per_cat}
We show per-category performance on 3D Average Precision (AP$_\threed$)  for \method{}-GEO and \method{}-LIFT in~\cref{tab:per_category_performance_geo,}.

\begin{table}[b]
\centering
\setlength{\tabcolsep}{15pt}
\begin{tabular}{l|rr}
\hline
Category &  GEO  &  LIFT \\ \hline
Board          & 12.42   &  9.92    \\ 
Printer        & 34.03  & 35.90   \\ 
Painting       & 5.16  &  6.31   \\ 
Microwave      & 33.47  &  48.54  \\ 
Tray           & 10.35  &  15.56   \\ 
Podium         & 41.74  &  62.18  \\ 
Cart           & 32.22  & 54.00  \\ 
Tram           & 0.25   & 8.65    \\ 
\midrule
\emph{Easy Categories} & 21.20   & 30.13  \\ 
\midrule
Monitor        & 18.25  & 13.92 \\ 
Bag            & 25.48   &  23.96  \\ 
Dresser        & 29.78  & 36.71   \\ 
Keyboard       & 15.44  & 12.58   \\ 
Drawers        & 30.01  & 57.21     \\ 
Computer       & 13.14  &  13.77   \\ 
Kitchen Pan    & 15.44   &  19.90  \\ 
Potted Plant   & 20.13   &  6.07  \\ 
Tissues        & 13.49   &   18.28   \\ 
Rack           &  14.60   &  15.74    \\ 
Toys           & 24.07   &  21.70   \\ 
Phone          & 22.21    &  11.37    \\ 
Soundsystem    & 17.73   &   17.69  \\ 
Fireplace      & 24.44  &  19.76     \\ 
\midrule
\emph{Hard Categories} & 20.30   & 20.62  \\ 
\midrule
\emph{All Categories} & 20.63  & 24.08  \\ 
\midrule
\end{tabular}
\caption{Per-category Performance of \method{}-GEO and \method{}-LIFT. The reported metric is AP$_\threed$ in target-aware evaluation. }
\label{tab:per_category_performance_geo}
\end{table}

\section{More Qualitative Results}
\label{sec:more_qualitative}

\begin{figure*}[t!]
    \centering
    \includegraphics[width=0.93\textwidth]{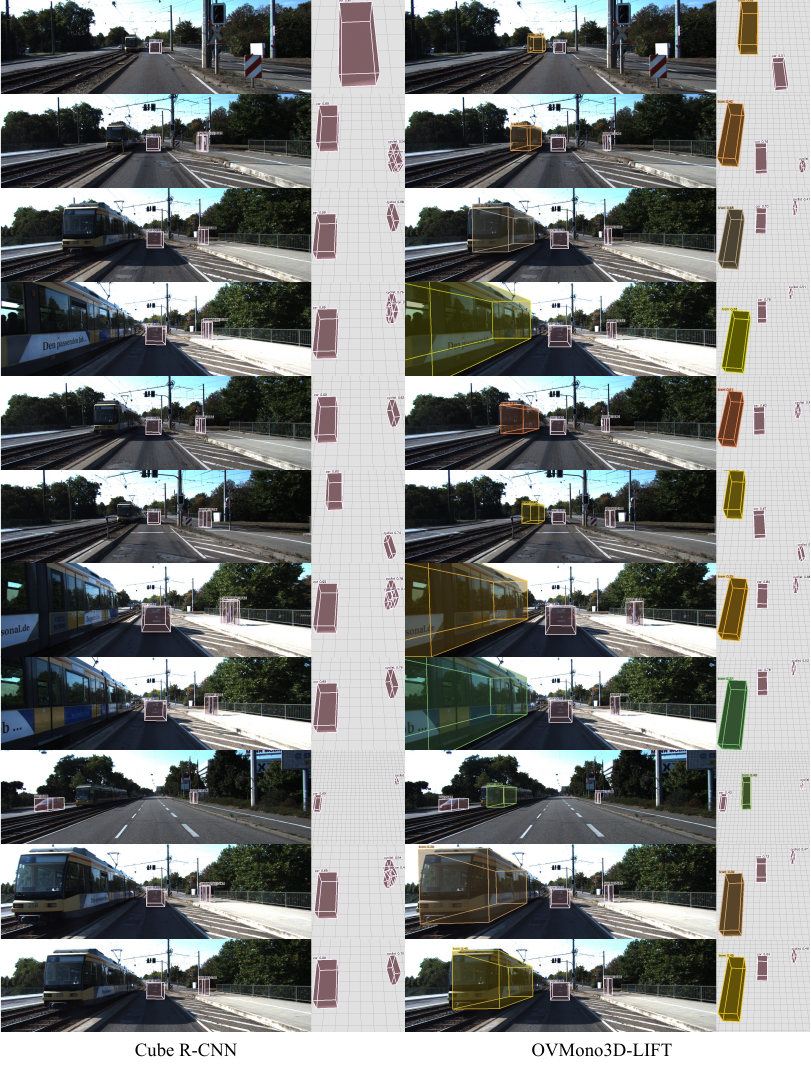}
    \vspace{-6mm}
    \caption{
\textbf{Qualitative Visualizations on the KITTI~\cite{geiger2012we} Test Set.} For each example, we present the predictions of Cube R-CNN~\cite{brazil2023omni3d} and \method{}-LIFT, displaying both the 3D predictions overlaid on the image and a top-down view with a base grid of $1\,\text{m} \times 1\,\text{m}$ tiles. Base categories are depicted with \textcolor{brown}{brown} cubes, while novel categories are represented in other colors.
    }
    \label{fig:outdoor_sup}
    
\end{figure*}

\begin{figure*}[t!]
    \centering
    \includegraphics[width=1\textwidth]{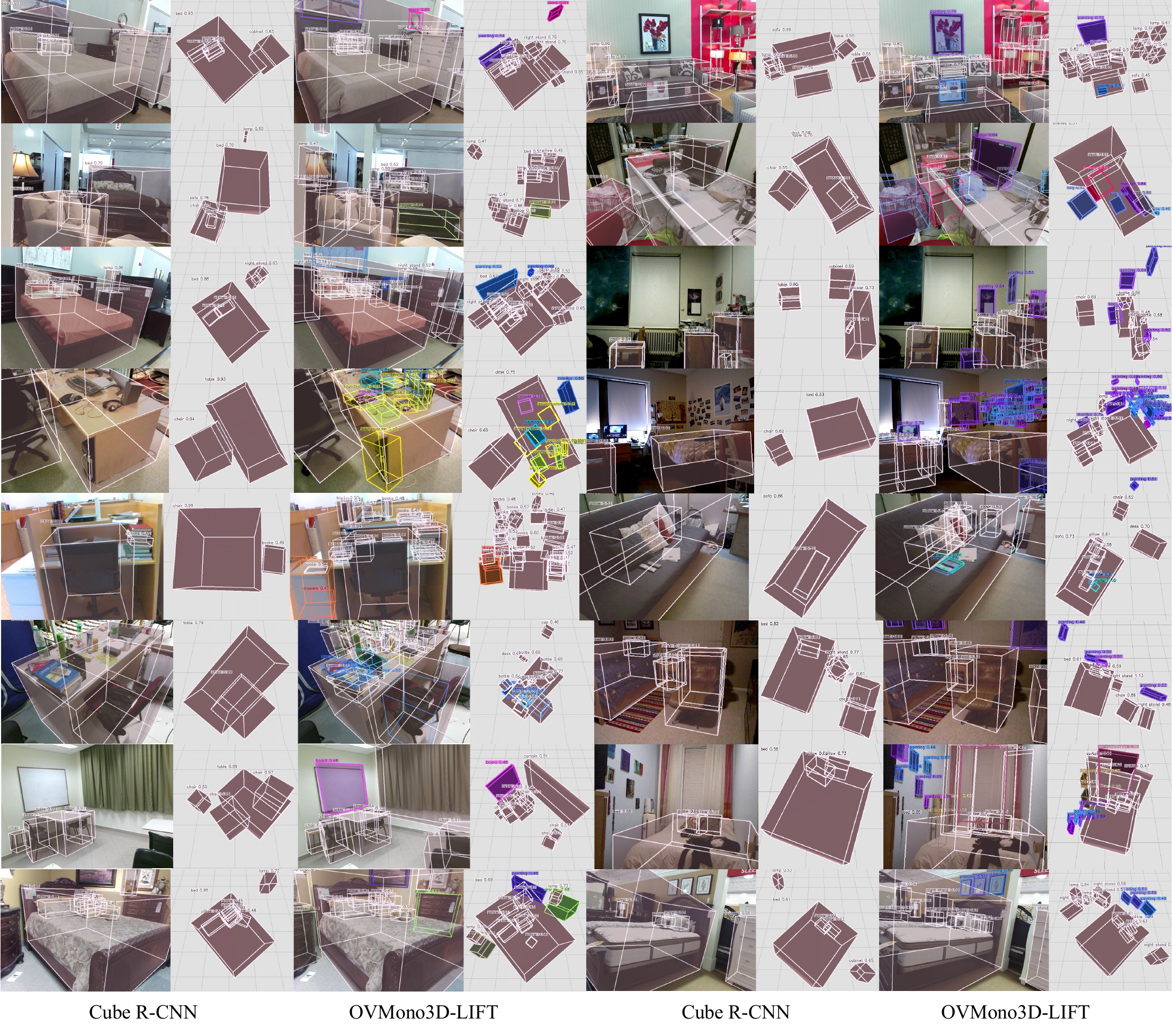}
    \caption{
\textbf{Qualitative Visualizations on the SUN RGB-D~\cite{song2015sun} Test Set.} For each example, we present the predictions of Cube R-CNN~\cite{brazil2023omni3d} and \method{}-LIFT, displaying both the 3D predictions overlaid on the image and a top-down view with a base grid of $1\,\text{m} \times 1\,\text{m}$ tiles. Base categories are depicted with \textcolor{brown}{brown} cubes, while novel categories are represented in other colors.
    }
    \label{fig:indoor_sup}
    
\end{figure*}

\begin{figure*}[t!]
    \centering
    \includegraphics[width=\textwidth]{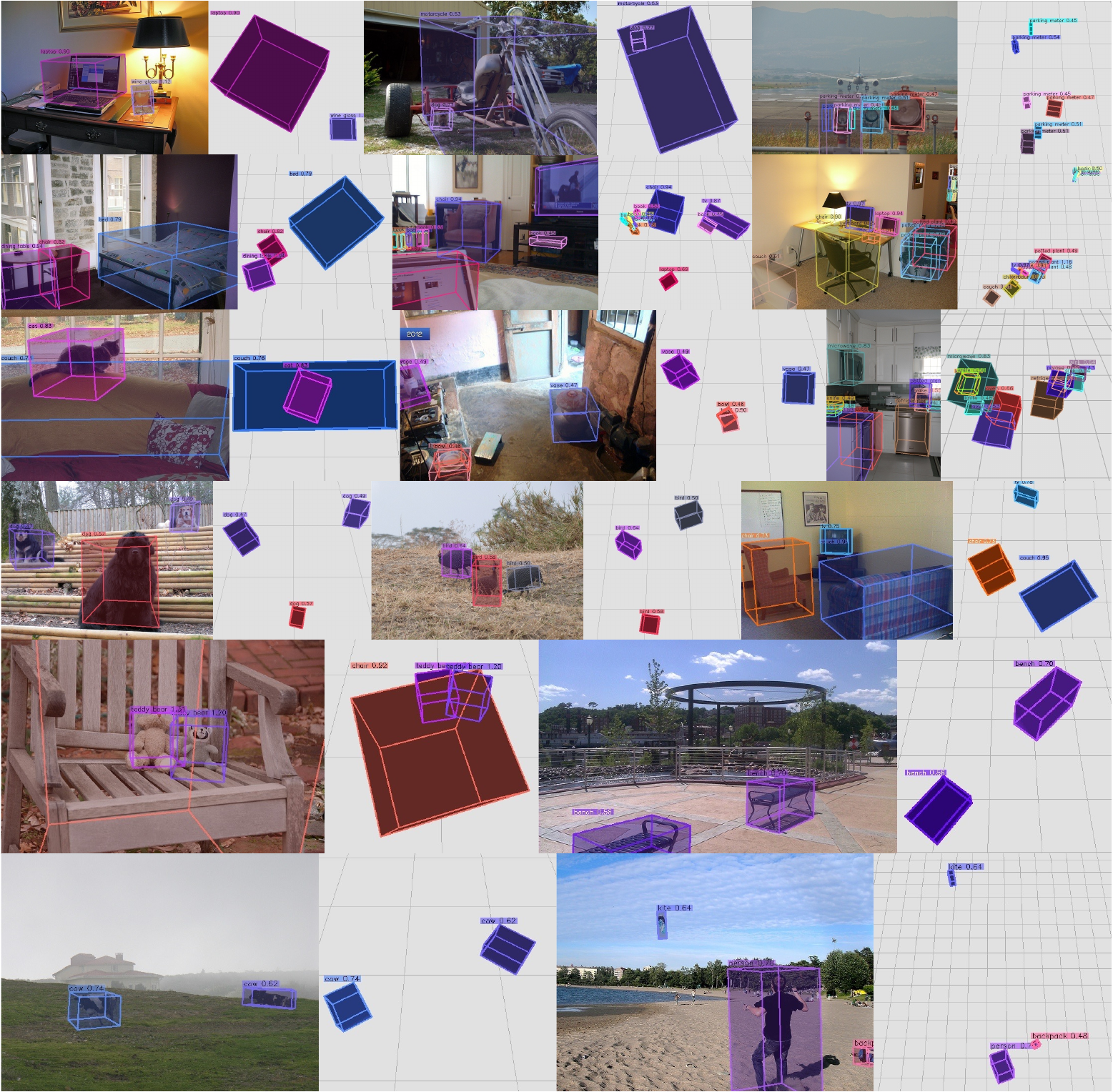}
    \caption{
\textbf{\method{}-LIFT on In-the-Wild COCO~\cite{lin2014microsoft} Images}. We display 3D predictions overlaid on the images and the top-down views with a base grid of $1\,\text{m} \times 1\,\text{m}$ tiles.
    }
    \label{fig:coco_sup}
    
\end{figure*}

Additional qualitative visualizations of \method{}-LIFT are provided for Omni3D~\cite{brazil2023omni3d} outdoor, indoor subsets, and COCO~\cite{lin2014microsoft} in-the-wild images in~\cref{fig:outdoor_sup,fig:indoor_sup,fig:coco_sup}, respectively. 
For COCO images, we visualize with intrinsics of $f = 2\cdot H$, $p_x = \frac{1}{2}W$, $p_y = \frac{1}{2}H$, where $H \times W$ is the input image resolution.

\begin{figure*}[t!]
    \centering
    \includegraphics[width=\textwidth]{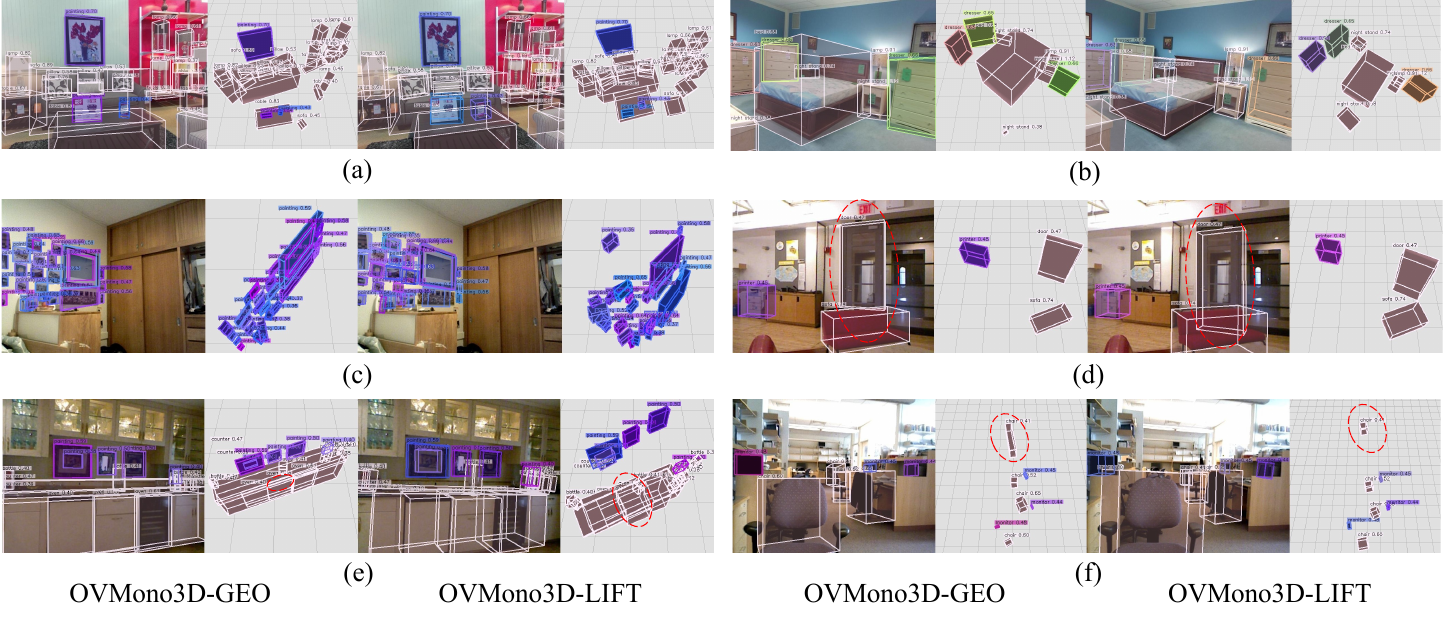}
    \caption{
\textbf{\method{}-GEO \textit{vs.} \method{}-LIFT on SUN RGB-D~\cite{song2015sun} Images}. For each example, we display the predictions of \method{}-GEO and \method{}-LIFT.  We display 3D predictions overlaid on the images and the top-down views with a base grid of $1\,\text{m} \times 1\,\text{m}$ tiles. Base categories are depicted with \textcolor{brown}{brown} cubes, while novel categories are represented in other colors.
    }
    \label{fig:geo_sup}
    
\end{figure*}

\cref{fig:geo_sup} illustrates a comparison between the predictions of \method{}-GEO and \method{}-LIFT.
\method{}-GEO derives object depth from an estimated metric depth map, yielding better relative depth consistency with scene layout (\eg, \cref{fig:geo_sup}c).
However, it estimates dimensions and poses based on visible object parts, leading to biases. For instance, it struggles with occlusions (\eg, the door in~\cref{fig:geo_sup}d), limited surface visibility (\eg, ovens in~\cref{fig:geo_sup}e), and noisy depth maps (\eg, the farthest chair in~\cref{fig:geo_sup}f). In contrast, \method{}-LIFT, leveraging learned priors, is more robust in such scenarios. Future work could integrate these methods to mitigate their respective limitations.

\section{Failure Cases}
\label{sec:fail}

\begin{figure*}[t!]
    \centering
    \includegraphics[width=\textwidth]{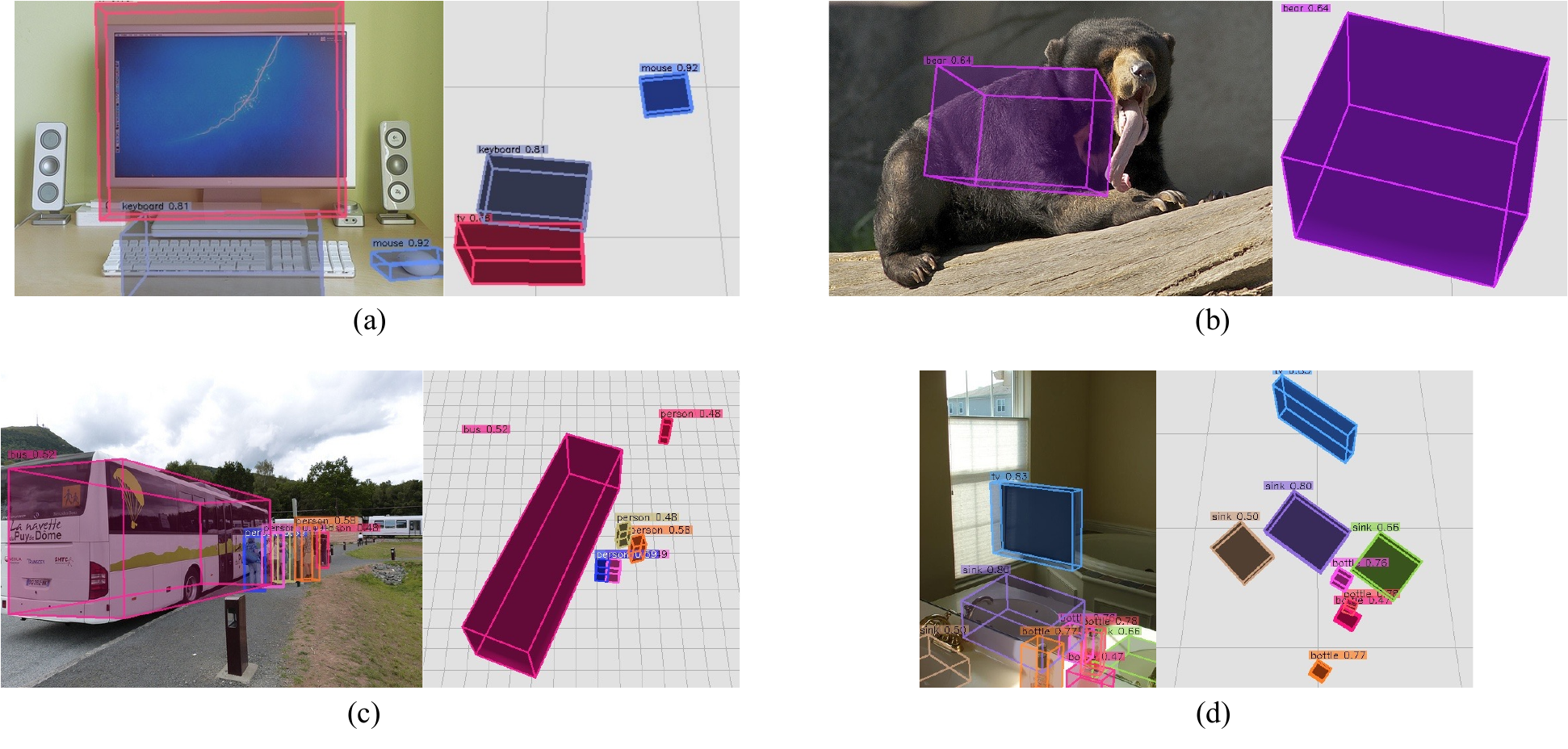}
    \caption{
\textbf{Failure Cases of \method{}-LIFT on COCO~\cite{lin2014microsoft} Images}. We display 3D predictions overlaid on the images and the top-down views with a base grid of $1\,\text{m} \times 1\,\text{m}$ tiles.
    }
    \label{fig:failure_sup}
    
\end{figure*}

\cref{fig:failure_sup} shows failure cases of \method{}-LIFT on COCO~\cite{lin2014microsoft} images. In~\cref{fig:failure_sup}a, the relative position from a top-down view is incorrect, indicating that our model sometimes predicts the wrong object depth.  In~\cref{fig:failure_sup}b, our model fails to predict the correct size and pose for the bear, suggesting that it may struggle with totally out-of-distribution objects. 
In~\cref{fig:failure_sup}c, our model fails to detect the person and bus in the distance, indicating that it may not perform well on small and distant objects. In~\cref{fig:failure_sup}d, our method fails to identify the mirror and incorrectly detects the object in the mirror. 
These failure cases suggest that our model still has room for improvement. Future research could explore better model architectures and weakly supervised learning techniques to address these shortcomings.

\section{More Analysis}
\label{sec:more_analysis}
\paragraph{Do synthetic data help?} 
We conducted an ablation study using synthetic data for \method{}-LIFT under resource-constrained conditions, with a frozen image encoder and excluded depth estimator. The synthetic data comes from Hypersim~\cite{roberts2021hypersim}, which provides indoor images rendered from artist-created meshes and serves as the sole synthetic data source in Omni3D~\cite{brazil2023omni3d}.

Tab.~\ref{tab:synthetic} presents the effect of synthetic data on the performance of \method{}-LIFT.
When synthetic data is incorporated alongside real data, a modest yet meaningful increase of 1 AP$_\threed$ point is observed in detecting objects from seen categories, while performance on novel categories remains largely unchanged. These findings suggest that, while synthetic data can enhance model performance in closed-vocabulary 3D object detection tasks, its benefits are minimal for detecting unseen objects, thereby limiting its usefulness in open-vocabulary 3D object detection scenarios.

\begin{table}[th!]
\centering
\begin{tabular}{@{}
   lccc}
\toprule
    Data & \#Images  & AP$_\threed^\text{Base}$ &  AP$_\threed^\text{Novel}$  \\ 
\midrule
Synthetic &55k & 7.14 &7.33 \\
Real &120k & 23.78 &\textbf{16.20}   \\
Synthetic+Real &175k & \textbf{24.77} &16.04  \\
\bottomrule
\end{tabular}
\caption{\textbf{Ablation on Synthetic Data.} 
Synthetic data refers to the Hypersim subset of the Omni3D dataset, while real data comprises the other Omni3D subsets.
Synthetic data boosts the performance of \method{}-LIFT on base categories but offers little benefit for novel objects.}
\label{tab:synthetic}
\vspace{-5mm}
\end{table}

\iffalse
\begin{table*}[!tp]
\centering
\setlength{\tabcolsep}{10pt}
\resizebox{0.85\textwidth}{!}{
\begin{tabular}{@{}
  l
  ccccccc}
\toprule
                  Feature Extractor&Arch. & Supervision & Dataset
                  & AP$_\threed$\textuparrow & AP$_\threed^\text{15}$\textuparrow & AP$_\threed^\text{25}$\textuparrow & AP$_\threed^\text{50}$\textuparrow    \\ 

\midrule
% MOCO-v2 & R50 & SSL & ImageNet1k\\
MAE~\cite{he2022masked}      &   ViT-B/16       &  SSL         &   ImageNet1k &7.72&11.95&6.43&0.80                     \\
CLIP~\cite{radford2021learning} & ViT-B/16 & VLM & WIT-400M  &9.02&14.19&8.44&0.50
\\
MiDas~\cite{ranftl2020towards} &   ViT-L/16       &  Depth         &   MIX-6  & 10.65&17.42&9.87&0.69
\\
% Cube R-CNN~\cite{brazil2023omni3d}          &  DLA34     & Detection &    Omni3D          & 11.21&15.59&10.13&1.32             \\
DINOv2~\cite{oquab2023dinov2} & ViT-B/14 & SSL & LVD-142M &\textbf{16.04} &\textbf{24.57} &\textbf{14.97} &\textbf{2.21}  
\\
\bottomrule
\end{tabular}
}
\caption{\textbf{Ablation Study on Feature Extractors for \method{}-LIFT in 3D Detection on Novel Categories.} We report AP$_\threed$ for 
 various feature extractors with frozen parameters during training.}
\label{tab:ablation_feature}
% \vspace{-2mm}
\end{table*}
\fi
\paragraph{Naming Ambiguity issue. } 
We quantitatively evaluate naming ambiguity in current 3D benchmarks using SUN RGB-D~\cite{song2015sun} as an example. For each object instance, we cropped its 2D bounding box and computed CLIP similarity scores between the visual features and all category text embeddings. We then aggregated these similarity vectors by ground-truth category and computed average similarities to form a confusion matrix, applying softmax normalization.

\begin{figure*}[t!]
    \centering
    \includegraphics[width=\textwidth]{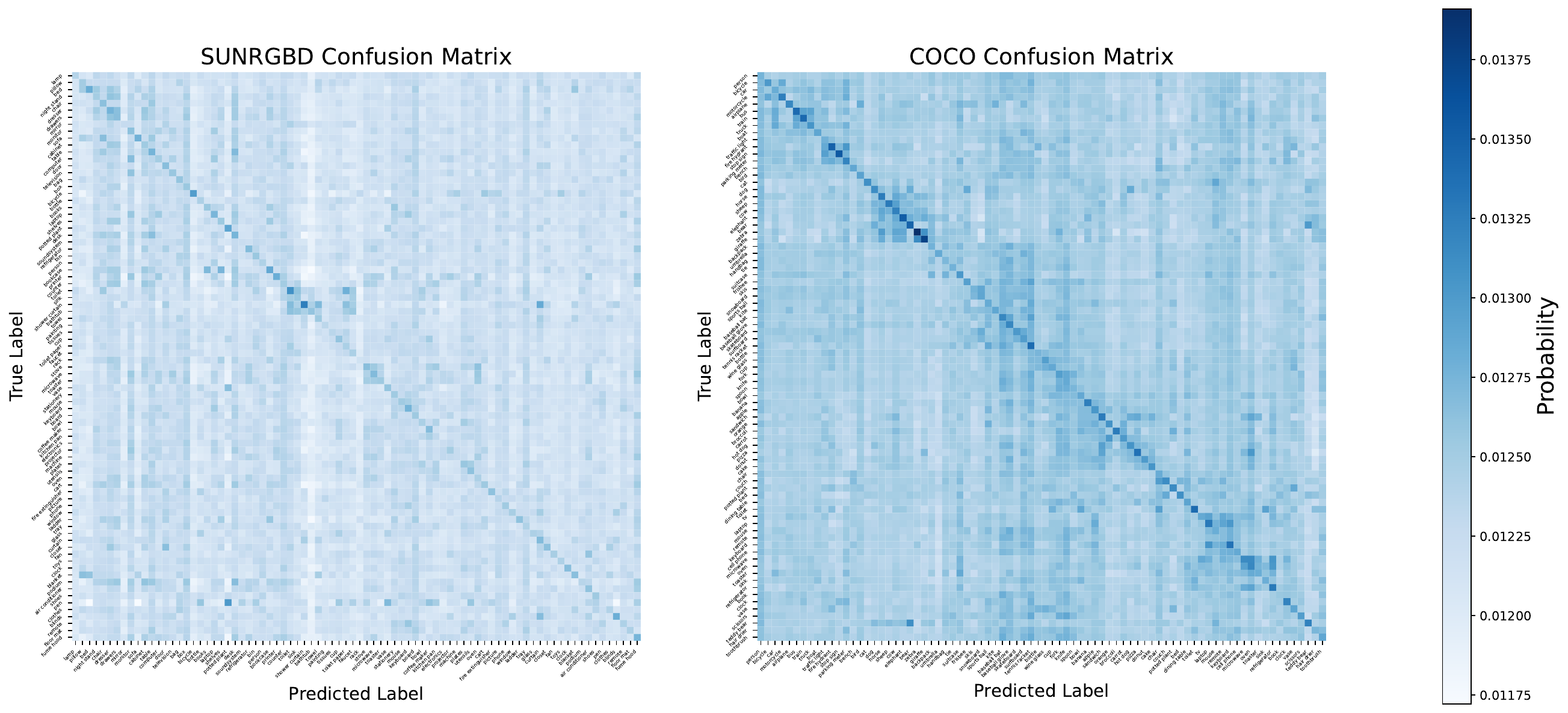}
    \caption{
Normalized confusion matrices displaying CLIP’s prediction performance on the SUNRGBD and COCO datasets.
    }
    \label{fig:naming_ambiguity}
    
\end{figure*}

As shown in~\cref{fig:naming_ambiguity}, SUN RGB-D annotations exhibit weaker self-correlation than COCO~\cite{lin2014microsoft}, indicating less distinct category boundaries. This reflects SUN RGB-D's highly similar category names (e.g., "table" vs. "desk"). In open-vocabulary settings, such similarity creates false negatives when models correctly identify a table as a desk—a distinction often acceptable in real-world applications. Therefore, our proposed target-aware evaluation is essential for datasets with ambiguous category definitions, unlike the well-differentiated categories in COCO.

\end{document}